\documentclass[lettersize,journal]{IEEEtran}
\usepackage{amsmath,amsfonts}
\usepackage{algorithmic}
\usepackage{algorithm}
\usepackage{array}
\usepackage[caption=false,font=normalsize,labelfont=sf,textfont=sf]{subfig}
\usepackage{textcomp}
\usepackage{stfloats}
\usepackage{xurl} 
\usepackage{verbatim}
\usepackage{graphicx}
\usepackage{cite}
\usepackage{romannum}
\usepackage{booktabs} 
\usepackage{tabularx} 
\usepackage{array} 
\usepackage{orcidlink} 
\usepackage{threeparttable}
\usepackage{pgfplots}

\newcommand\submittedtext{%
  \footnotesize This work has been submitted to the IEEE Transactions on Affective Computing for possible publication. Copyright may be transferred without notice, after which this version may no longer be accessible.}

\newcommand\submittednotice{%
\begin{tikzpicture}[remember picture,overlay]
\node[anchor=south,yshift=10pt] at (current page.south) {\fbox{\parbox{\dimexpr0.65\textwidth-\fboxsep-\fboxrule\relax}{\submittedtext}}};
\end{tikzpicture}%
}
\pgfplotsset{compat=1.18}

\begin{document}

\title{Spoken in jest, detected in earnest: A systematic review of sarcasm recognition — multimodal fusion, challenges, and future prospects}

\author{Xiyuan Gao\,\orcidlink{0000-0003-0870-6721},  Shekhar Nayak\,\orcidlink{0000-0002-4277-4851}, Matt Coler\,\orcidlink{0000-0002-7631-5063}

\thanks{Manuscript submitted on November 19, 2024; revised on May 24, 2025. \textit{(Corresponding author: Xiyuan Gao)}}
\thanks{X.Gao, S.Nayak and M.Coler are with Campus Fryslân, University of Groningen, Leeuwarden 8911 CE, the Netherlands (e-mails: xiyuan.gao@rug.nl; s.nayak@rug.nl; m.coler@rug.nl)}
}

\maketitle
\submittednotice

\begin{abstract}
Sarcasm, a common feature of human communication, poses challenges in interpersonal interactions and human-machine interactions. Linguistic research has highlighted the importance of prosodic cues, such as variations in pitch, speaking rate, and intonation, in conveying sarcastic intent. Although previous work has focused on text-based sarcasm detection, the role of speech data in recognizing sarcasm has been underexplored. Recent advancements in speech technology emphasize the growing importance of leveraging speech data for automatic sarcasm recognition, which can enhance social interactions for individuals with neurodegenerative conditions and improve machine understanding of complex human language use, leading to more nuanced interactions. This systematic review is the first to focus on speech-based sarcasm recognition, charting the evolution from unimodal to multimodal approaches. It covers datasets, feature extraction, and classification methods, and aims to bridge gaps across diverse research domains. The findings include limitations in datasets for sarcasm recognition in speech, the evolution of feature extraction techniques from traditional acoustic features to deep learning-based representations, and the progression of classification methods from unimodal approaches to multimodal fusion techniques. In so doing, we identify the need for greater emphasis on cross-cultural and multilingual sarcasm recognition, as well as the importance of addressing sarcasm as a multimodal phenomenon, rather than a text-based challenge.
\end{abstract}


\section{Introduction}
\IEEEPARstart
{S}arcasm is a ubiquitous feature of everyday conversations, where receiving compliments or praise from friends can often leave us questioning their sincerity. This moment of doubt emerges when we detect subtle cues beyond the spoken words, such as a wink or a trace of sarcasm in the melody. Adding to its complexity, sarcasm does not always function as a mere polarity switcher; it can convey messages that deviate from the literal content \cite{Sperber}. It operates subtly yet remarkably effective in a social context. Researchers have highlighted its multifaceted role. Jorgensen \cite{Jorgensen} and Brown \cite{Brown} mentioned sarcasm is applied to address complaints and criticism to close relationships in a less harmful way. Seckman and Couch \cite{Seckman} claimed that sarcasm fortifies the solidarity within work groups. Perceiving sarcasm, however, demands more cognitive effort than understanding direct expressions \cite{Giora,McDonald}, primarily because sarcasm transcends the literal content of the spoken words \cite{Grice}, and relies on pragmatic cues, such as information from preceding discourse, shared knowledge \cite{Jorgensen1984}, perceptual indicators, linguistic signals, societal norms, and even the speaker’s preferences or background \cite{McDonald}. The accurate interpretation of sarcasm is vital for effective social interaction, yet it poses significant challenges for individuals with neurodegenerative conditions, such as semantic dementia, or for those on the autism spectrum \cite{Kipps,Persicke} 

The importance of speech prosody in sarcasm perception has been validated by linguists across languages. Previous research in linguistics related fields has predominantly investigated how sarcasm is conveyed through various sound patterns in language, highlighting the intricate relationship between speech delivery and the intended sarcastic meaning. For instance, Rockwell \cite{Rockwell} observed that in English, sarcasm is often conveyed through a slower tempo, increased intensity, and a lower pitch. Similarly, Cheang and Pell \cite{Cheang_2009} found that sarcastic speech in Cantonese is characterized by an elevated pitch, reduced intensity, a slower rate, and less vocal noise. Scharrer and Christmann \cite{Scharrer} identified sarcasm in German by features such as a lower pitch, heightened intensity, and extended vowel duration. Loevenbruck \textit{et al}. \cite{Lœvenbruck} revealed that sarcastic expressions are marked by longer duration, as well as increased pitch level and range. These findings collectively emphasize the nuanced role of prosodic features in signaling sarcasm across different languages.

Early efforts in recognizing sarcasm through computational models primarily focused on leveraging speech data for potential applications in dialogue systems \cite{Tepperman,Rakov}. Sequentially, researchers have delved into the development of models capable of recognizing sarcasm in text over the past decades. Sarcasm text expressions are often described as “noisy” data, introducing ambiguity that can disrupt the accuracy of language models. This poses a significant challenge for tasks like Sentiment Analysis (SA), which are heavily influenced by the presence of sarcasm. Review works in computational processing of sarcasm have primarily concentrated on text data sourced from social networks \cite{Joshi,Eke,Chaudhari}. 

With the rapid development of speech technology and spreading use of voice-assisted devices, such as Alexa and Google Assistant, it is increasingly important to acknowledge the role of speech data in sarcasm recognition. Sarcasm recognition in speech holds the potential to assist individuals with neurodegenerative conditions by enhancing their ability to navigate and integrate into social conversations. Furthermore, in the realm of Human-Machine Interaction (HMI), the capability to accurately detect sarcasm enables machines to better grasp the subtleties of human communication, thereby fostering more nuanced and effective interactions. In light of this, this paper presents a comprehensive systematic review of sarcasm recognition, with a dedicated focus on speech data, presenting the development from unimodal to multimodal approaches, aiming to fill the vacancy of a systematic review on such a topic. In detail, we inspect three aspects including datasets, features, classification methods. Through the systematic review, we answer the following research questions (RQs):

\noindent \textbf{RQ1:} \textit{What are the available datasets for sarcasm recognition using speech data, and what limitations do they present? What guidelines should be followed for creating high-quality datasets in this field?}\\
\noindent \textbf{RQ2:} \textit{As the field progresses from unimodal to multimodal sarcasm recognition, how have feature extraction developed? What are the challenges and limitations in current feature extraction methods?}\\
\noindent \textbf{RQ3:} \textit{As the field progresses from unimodal to multimodal sarcasm recognition, how have classification methods evolved? What are the constraints of current classification methods?}\\
\noindent \textbf{RQ4:} \textit{What are the primary challenges and limitations in the current development of sarcasm recognition technology, and what are the practical implications in real-world scenarios?}

This is the first systematic review of sarcasm recognition that focuses on speech data, traversing from unimodal to multimodal approaches. We aim to be a valuable resource for researchers and practitioners, offering insights into the latest developments and discussing challenges for future growth. This review serves as a bridge connecting various related research domains. Our goal is to inspire collaborative efforts in finding collective solutions for this multifaceted challenge.

The structure of this paper is as follows: Section \Romannum{2} discusses the methods applied to operate the systematic review. Section \Romannum{3} summarizes the results and analyzes accordingly to address the research questions. Section \Romannum{4} discusses the results of this systematic review and points out the limitations of the current research. Finally, Section \Romannum{5} provides the conclusion of the review.

\section{Methods}
This paper uses systematic review which follows the guidelines suggested by PRISMA \cite{Moher}. This systematic review follows the PRISMA checklist \footnote{\scriptsize\url{https://www.prisma-statement.org/prisma-2020-checklist}}. Items related to risk of bias (\#12, 15, 19, 22), effect measures (\#13), additional analyses (\#16, 23) are excluded as they are beyond the scope of this systematic review. 

To gather articles addressing our research questions, we selected the following databases: Scopus, IEEE Xplore, ISCA archive, ScienceDirect, and ACM Digital Library. These databases were chosen because they contain large amounts of peer-reviewed articles published in the field of computer science. 
Each database was initially searched between 1st Aug. 2023 and 5th Aug. 2023. To ensure the currency of our review and in response to reviewer feedback, we conducted a follow-up search on 24 Mar. 2025, covering publications up to Dec. 2024. The same methodology and search strings were applied to maintain consistency with the initial search protocol.

\begin{enumerate}
  \item To define the methodology involved in the articles, we applied \textit{automatic, machine learning, deep-learning}; 
  \item To include articles that availed a multimodal approach, we used \textit{multimodal, multi-modal, multimodality}; 
  \item To allocate the main phenomenon we concern, we included \textit{sarcasm, irony, detection, recognition}; 
  \item To define the scope of the applied data type, we indicated \textit{speech, audio, voice, sound, conversation, dialogue}. 
\end{enumerate}

These terms above were chosen based on their prevalence in the preliminary search. To ensure alignment with the search logic employed by the chosen databases, we further combined these terms as follows in our practical application:

\begin{itemize}
  \item \textit{(``sarcasm'' OR ``irony'') AND (``detection'' OR ``recognition'') AND (``speech'' OR ``audio'' OR ``voice'' OR ``sound'' OR ``conversation'')}
  \item \textit{(``multimodal'') AND (``sarcasm'' OR ``irony'') AND (``detection'' OR ``recognition'') AND (``speech'' OR ``audio'' OR ``voice'' OR ``sound'' OR ``conversation'' OR ``dialogue'')}
  \item \textit{(``machine learning'' OR ``deep learning'' OR ``automatic'') AND (``sarcasm'' OR ``irony'') AND (``detection'' OR ``recognition'') AND (``speech'' OR ``audio'' OR ``voice'' OR ``sound'' OR ``conversation'' OR ``dialogue'')}
\end{itemize}

When conducting search in the mentioned databases, we applied a few search limits to exclude articles beyond our research scope. First, we limited the research domain to computer science. Second, the article type was defined to be empirical study. Third, only articles written in English were included. 
Fourth, the selected articles should be published. Initially, we limited the publication window to the period from 2006 to 2023. In the follow-up search in Mar. 2025, we extended the publication window to include articles published in 2024. Due to database constraints that allow filtering by year but not by month, our follow-up search included the full year of 2023, which overlapped with the initial search conducted in Aug. 2023. To address this, we combined the records from both searches and conducted deduplication prior to screening.

To gather the eligible articles, a number of inclusion and exclusion criteria were defined for both the abstract and full-text screening processes: 

\begin{enumerate}
    \item The selected papers are in the computer science domain.
    \item The selected papers are empirical study instead of review, report, or commentaries.
    \item The selected papers contain sarcasm detection/recognition in speech data.
    \item The selected papers include a machine learning based method.
    \item The selected papers are written in English.
    \item Any unrelated, duplicated, unavailable full-texts, abstract-only papers will be removed.
\end{enumerate}

From the selected articles, we conducted a structured data analysis process. First, we stored the articles in Zotero for easy access. Second, to answer the research questions, we created a data extraction form in a spreadsheet. This form included information (e.g., dataset language, data size, annotation process, etc.) related to the research questions, and as we reviewed the selected papers, we recorded the pertinent information in the corresponding cells. This meticulous data extraction process ensured that all research questions were adequately addressed. In terms of data analysis, we access this from three perspectives: data, features extraction, and classification. These perspectives reflect the major components of the entire process of sarcasm recognition. 

\begin{figure}
	\centering
\includegraphics[width=0.50\textwidth]{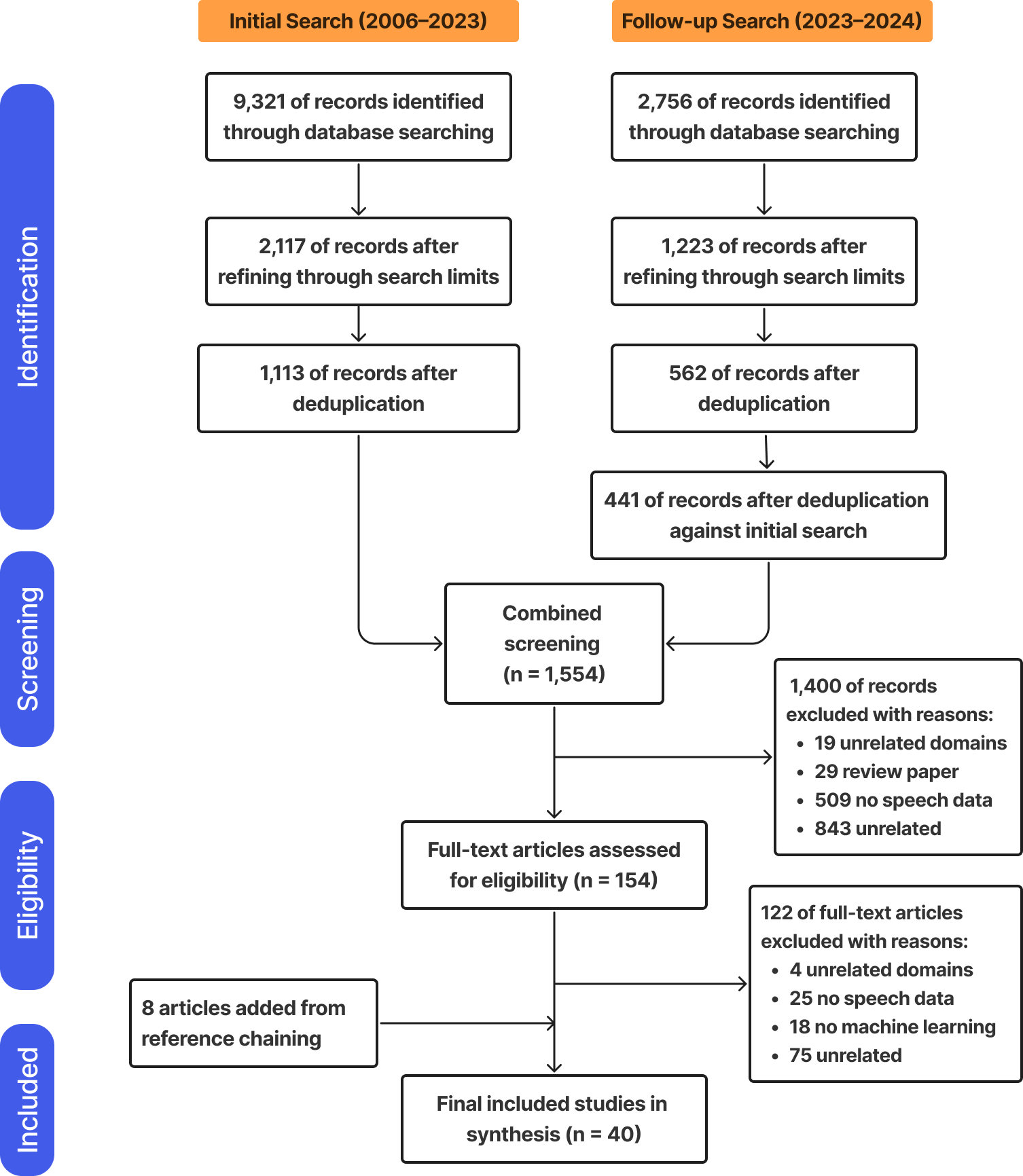}
\caption{PRISMA-guided systematic review process. The initial search phase covers articles from 2006 to Aug. 2023 and the follow-up search phase covers articles from Aug. 2023 through Dec. 2024. The search results from both phases are deduplicated prior to screening.}
	\label{filtering}
\end{figure}

\section{Findings and analysis}
A complete screening and selection process is illustrated in Figure \ref{filtering}. During the initial search phase, a total of 9,321 records were collected using the predefined search strings. After applying search limits across selected databases, 2,117 records remained. Following deduplication, 1,113 articles were retained and imported into Zotero for screening. We applied a two-stage screening process: (1) title and abstract screening using defined inclusion and exclusion criteria, followed by (2) full-text screening based on the same criteria. Along with this full-text screening, we also investigated the related works and references of these articles to extract additional qualified articles. It resulted in 21 articles from the full-text screening and 7 more from related work references, yielding 28 studies in total. To ensure comprehensive coverage of recent literature, we conducted a follow-up search phase on 24th Mar. 2025, collecting articles from 2023 to 2024. This search identified 2,756 records, from which 1,223 remained after applying search limits. After conducting cross-phase deduplication, 441 unique records remained. At last, 11 articles were gathered after full-text screening and one was added via reference chaining, creating an additional 12 studies. In total, 40 articles were included in the final synthesis. 

\begin{figure}[t]
	\centering
\includegraphics[width=0.4\textwidth]{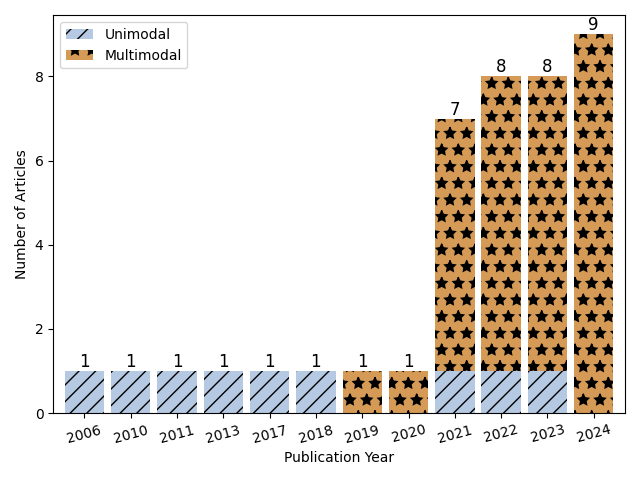}
\caption{Trend of the publications. ``Multimodal'' approaches include audio with text, audio with articulatory features, or audio with both text and visual modalities, while ``Unimodal'' refers to audio-only approaches.}
	\label{figurepub}
\end{figure}

Figure~\ref{figurepub} offers an overview of the publication years of the selected articles. A noteworthy trend emerges: prior to 2019, there were sporadic publications related to sarcasm recognition in speech. However, an increase is evident after 2019, with the majority of recent articles adopting a multimodal approach. Except for the emerging research interest in multimodality, this surge in research can be attributed to the release of the MUStARD dataset \cite{Castro} in 2019, which catalyzed interest and innovation in multimodal sarcasm recognition. The availability of such resources has fueled new research directions in the field. 

\begin{figure}[h]
    \centering
    \includegraphics[width=0.3\textwidth]{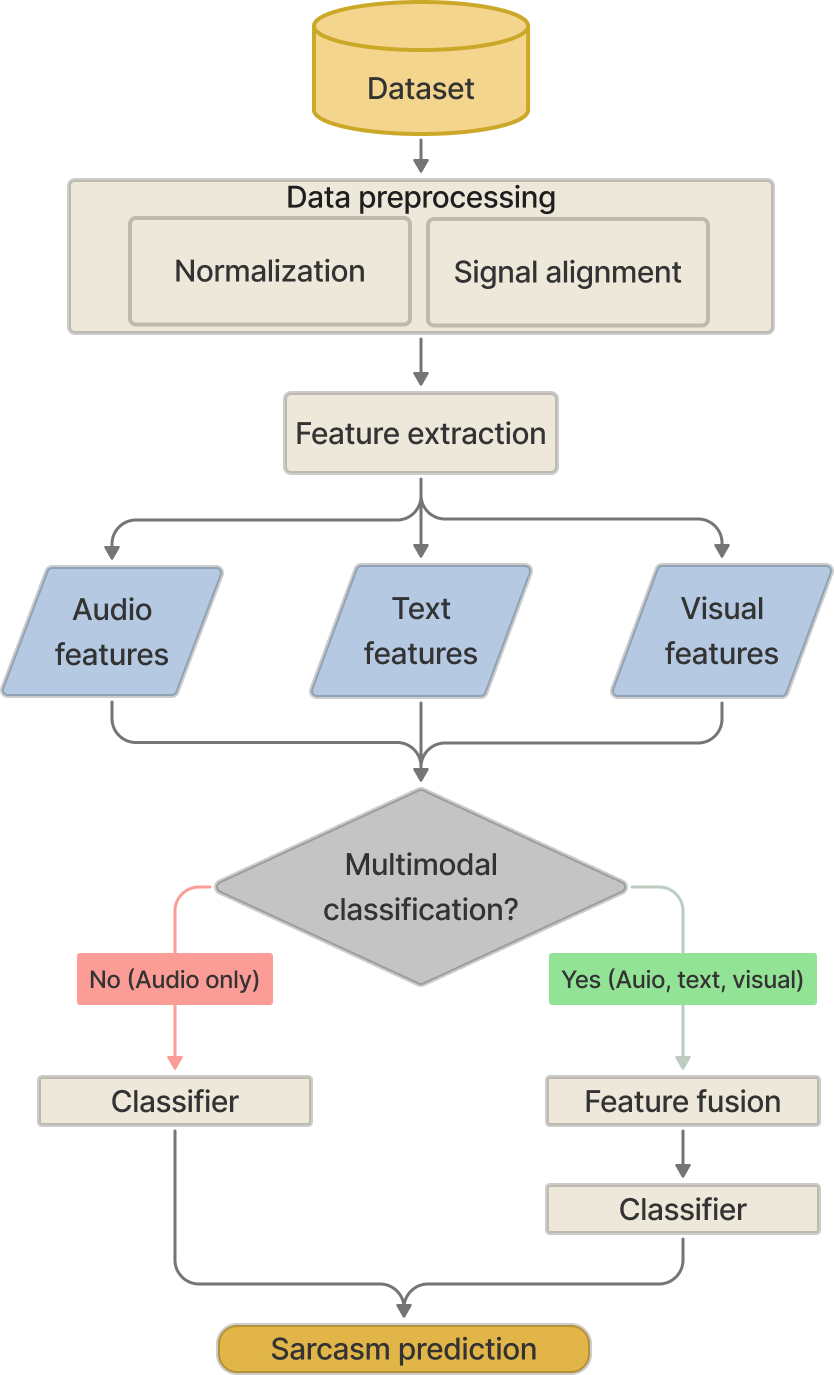}
    \caption{The overall process of sarcasm recognition.}
    \label{overview}
\end{figure}
To provide a unified view of the overall process of sarcasm recognition, Figure \ref{overview} illustrates a general pipeline. The pipeline begins with a multimodal dataset that typically comprises audio, textual, and visual signals. These inputs undergo preprocessing steps such as normalization, noise removal, and signal alignment. Next, modality-specific features are extracted, for example, prosodic cues like pitch and duration from audio; lexical and contextual embeddings from text; and visual cues such as eye gaze and facial expressions. The system then proceeds to classification via two distinct paths: unimodal and multimodal. With a special focus on speech, the unimodal classification relies on features from the audio modality, employing traditional machine learning or deep learning models. In contrast, multimodal classification integrates features across modalities through appropriate fusion strategies before passing them to a final decision layer. The final output is typically a binary label indicating whether the input instance is sarcastic or not. 
In the following content, we follow the pattern of datasets, feature extraction, and classification to present the analysed results. 

\subsection{Dataset for sarcasm recognition}
In addressing RQ1, we conducted a comprehensive examination of the existing datasets utilized for sarcasm recognition, presenting key features, such as data source, modality, labeling, and more. Through this comparative assessment, we are able to elucidate the respective strengths and weaknesses inherent in these datasets. In our review, two distinct categories of datasets emerged: those that are curated to cater to a wider user base, and those that are designed to serve the needs of a particular research endeavor. As the latter resource is not available for open-source use or public access, the eligibility is inaccessible; we focus on the former category of open-source datasets in this review. Further, we delved into the development processes of these datasets, dissecting their unique characteristics. This exploration served to shed light on the processes involved in creating high-quality datasets for sarcasm recognition.

\subsubsection{Characteristics of existing datasets}
An overview of the prominent datasets is presented in Table~\ref{tabledata}, offering a comprehensive view of their key attributes. 
The entries are listed in ascending order, organized according to the year of publication. Our review of the datasets revealed the following key characteristics:

\begin{table*}[t!]
\centering
\begin{threeparttable}
\footnotesize
\caption{Datasets for sarcasm recognition in speech. \label{tabledata}}
\begin{tabularx}{\textwidth}{
>{\raggedright\arraybackslash}p{1.3cm} >{\raggedright\arraybackslash}p{0.7cm} >{\raggedright\arraybackslash}p{1.5cm} >{\raggedright\arraybackslash}p{1cm} >{\raggedright\arraybackslash}p{1cm} >{\raggedright\arraybackslash}p{6.8cm} >{\raggedright\arraybackslash}p{0.7cm} >{\raggedright\arraybackslash}p{0.7cm} >{\raggedright\arraybackslash}p{1cm}} 
\toprule
\textbf{Dataset} & \textbf{Year} & \textbf{Source} & \textbf{Language} & \textbf{Modality} & \textbf{Labels} & \textbf{Context} & \textbf{Speaker} & \textbf{Size (h)} \\ 
\midrule
\href{https://cse.iitkgp.ac.in/~ksrao/res.html}{IITKGP-SEHSC} & 2011 & Recordings & Hindi & A & Anger, disgust, fear, happiness, sadness, neutrality, sarcasm, surprise & No & No & 7 \\
\href{https://github.com/soujanyaporia/MUStARD}{MUStARD} & 2019 & TV series & English & T, A, V\tnote{a} & Sarcasm, non-sarcasm & Yes & Yes & 3.68 \\
\href{https://github.com/GussailRaat/ACL2020-SE-MUStARD-for-Multimodal-Sarcasm-Detection}{SEmoji MUStARD} & 2020 & TV series (MUStARD) & English & T, A, V & Sarcasm, non-sarcasm, implicit/explicit sentiment, disgust, happy, surprised, neutral, anger & Yes & Yes & 3.68 \\
\href{https://zenodo.org/record/4707913}{Spanish Dataset} & 2021 & Animations & Spanish & T, A, V & Sarcasm, non-sarcasm, sarcasm sub-types & No & Yes & - \\
\href{https://docs.google.com/forms/d/e/1FAIpQLScEFtQ2mCkmaISf5w4YXjIoBV8W-1wISSW8r1WbnIPIuW40iA/viewform}{SEEmoji MUStARD} & 2022 & TV series (MUStARD) & English & T, A, V & Sarcasm, non-sarcasm, implicit/explicit sentiment, disgust, happy, surprised, neutral, anger, emojis & Yes & Yes & 3.68 \\
\href{https://github.com/helengent/Irony-Recognition}{Irony-Recognition} & 2022 & Podcast & English & A & Sarcasm, non-sarcasm & No & Yes & 4.68 \\
\href{https://github.com/apoorva-nunna/MUStARD_Plus_Plus}{MUStARD++} & 2022 & TV series (MUStARD, MELD) & English & T, A, V & Sarcasm, sarcasm sub-types, non-sarcasm, implicit/explicit sentiments, valence and arousal rating, excitement, fear, sad, frustrated, disgust, happy, surprised, ridicule, neutral, anger & Yes & Yes & 7.36 \\
\href{https://github.com/LCS2-IIITD/MSH-COMICS}{MaSaC} & 2023 & TV series & Hindi-English & T, A, V & Sarcasm, non-sarcasm, humorous, non-humorous & No & Yes & - \\
\href{https://github.com/annoymity2022/Chinese-Dataset}{CMMA} & 2023 & TV series & Chinese & T, A, V & Sarcasm, non-sarcasm, humorous, non-humorous, sentiments (positive, negative, neutral), emotions (joy, sadness, anger, fear, surprise, disgust, neutral), pride, love, sentiment-emotion and sarcasm-humor inter-relatedness measures & Yes & Yes & 15.2 \\
\bottomrule
\end{tabularx}
\begin{tablenotes}
\scriptsize
\item[a] T indicates text, A indicates audio, and V indicates video.
\end{tablenotes}
\end{threeparttable}
\end{table*}

\begin{figure}[h]
	\centering
\includegraphics[width=0.45\textwidth]{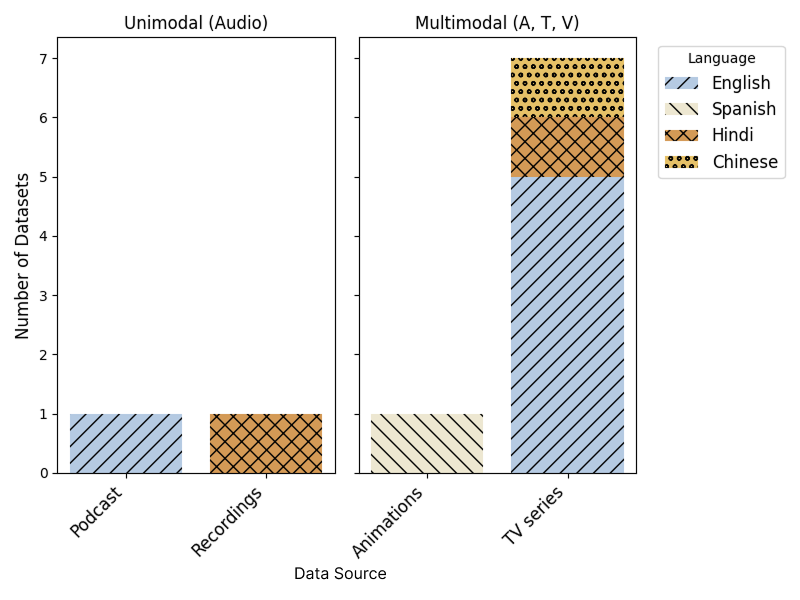}
\caption{Distribution of datasets by source, language, and modality.}
	\label{data_source}
\end{figure}

\paragraph{Source} In our review of sarcasm recognition datasets containing speech data, we identified four primary data sources: recordings, TV series, animations, and podcasts. 
As shown in Figure \ref{data_source}, most of the surveyed studies use TV series as a data source. This is due to two major merits of the TV-sourced data. First, compared to acted data, TV series ought to be more natural as the actors are performing a real-life scenario. Second, the collection cost is comparatively low compared to recording in a studio environment.

\paragraph{Language} As shown in Figure \ref{data_source}, the language diversity within these datasets is limited, with the exception of IITKGP-SEHSC (Hindi) \cite{Koolagudi}, MaSaC (Hindi and English) \cite{Bedi}, Spanish Dataset (Spanish) \cite{Alnajjar}, and CMMA (Chinese) \cite{Zhang_CMMA}. The majority of the datasets are available exclusively in English.

\paragraph{Modality} IITKGP-SEHSC and Irony-Recognition \cite{Gent} exclusively contain audio data, while the remaining datasets are meticulously curated to facilitate multimodal (text, audio, visual) sarcasm recognition. 

\paragraph{Labeling} SEmoji MUStARD \cite{Chauhan_2020} and SEEmoji MUStARD \cite{Chauhan_2022} are extensions of the original MUStARD. MUStARD++ \cite{Ray} are sourced from both MUStARD and MELD \cite{Poria}. The extensions primarily focus on enhancing the labels, transitioning from a mere binary classification (i.e., sarcasm and non-sarcasm) to a more comprehensive representation of associated sentiments and emotions. Beyond assigning affective labels, CMMA provides labels capturing the relevance between sentiment and emotion, and between sarcasm and humor. These labels are represented on a five-level scale: [-2, -1, 0, 1, 2], indicating negative, neutral, and positive relevance. This enables informed selection of main and auxiliary tasks in multi-task learning models.

\paragraph{Context} Context is defined as the preceding dialogue turns of a labeled utterance. Given the pragmatic nature of sarcasm, linguists have posited that context plays a crucial role in reducing ambiguity in sarcasm interpretation \cite{Riviere,Utsumi,KumonNakamura}. Consistent with this view, several datasets collected for sarcasm research include contextual information along with the labeled utterances \cite{Castro,Chauhan_2020,Chauhan_2022,Ray, Zhang_CMMA}. 

\paragraph{Speaker information} In the aspect of sarcasm perception, Fan \textit{et al}. \cite{Fan_2015} found that anticipation of sarcastic intent is crucial for the efficient comprehension of sarcasm. That is to say, when sarcasm is a recurring style of a particular speaker, the likelihood of sarcastic expression from that individual increases, thereby speaker information significantly shapes the interpretation of the statements. In our review, speaker information has been leveraged by researchers \cite{Castro,Chauhan_2020,Chauhan_2022,Ray,Wu} to establish two evaluation settings: speaker-dependent and speaker-independent. In the speaker-dependent setting, utterances from the same speakers are used in both the training and testing sets, whereas the speaker-independent setting involves different speakers in the training and testing sets.

\paragraph{Size} 
Sarcasm datasets are typically limited in size, with only one large dataset encompassing 15.2 hours of video that includes both sarcastic and non-sarcastic instances \cite{Zhang_CMMA}. This constraint reflects the challenge in the annotation sarcasm, particularly in achieving consensus among different annotators. Despite this challenge, there is a noticeable trend of gradual expansion in the size of these datasets over years, reflecting ongoing efforts to accumulate more comprehensive data sets for sarcasm research.

\subsubsection{Dataset development}
The creation of a sarcasm dataset typically involves the following stages: data collection, data preprocess and data annotation. Each of them plays a crucial role in shaping the quality and usability of the resultant dataset.

\paragraph{Data collection} This initial phase involves the acquisition of raw data from a diverse array of sources. In our findings, examples of acted speech datasets, such as IITKGP-SEHSC and the dataset created by Geng \textit{et al}. \cite{Geng}, involve trained vocal professionals who generate utterances, based on carefully designed stimuli. Notably, our review highlights a growing trend of utilizing TV series as a source of acted speech data. Although TV series involve acted performances, the actors often strive to replicate real-life scenarios, resulting in datasets that tend to exhibit a more natural conversational tone. Another type of data source is used by Gent \textit{et al}. \cite{Gent}, they collected 4.68 hours of speech data from a podcast. Innovatively, Burkhardt \textit{et al}. \cite{Burkhardt} built an interactive mobile phone-based interface to collect sarcastic speech. Users were guided to record their sarcastic reaction to visual stimuli and labeled their speech sequentially. 

\paragraph{Data preprocessing} The data preprocessing stage is necessary for enhancing data quality and consistency. It encompasses various tasks, such as format conversion, audio intensity normalization \cite{Atassi_2010}, segmentation \cite{Castro,Arun}, and signal alignment \cite{Alnajjar,Bedi}. Signal alignment, in particular, holds significant importance when working with multimodal data. However, it’s also frequently seen that cross-modality alignment is implemented as part of the modeling process. For instance, Wu \textit{et al}. \cite{Wu} used GENTLE \footnote{\url{https://rmozone.com/gentle/}} to align video and audio after extracting the related features. Hasan \textit{et al}. \cite{Hasan} applied the P2FA forced alignment \footnote{\url{https://babel.ling.upenn.edu/phonetics/old_website_2015/p2fa/index.html}} to extract the timing of key words and trace the aligned visual and acoustic features. 
In multimodal analysis, transcription is often needed. Zhang \textit{et al}. \cite{Zhang_CMMA} utilized the Google Cloud Speech-to-Text service\footnote{\url{https://cloud.google.com/speech-to-text}} to transcribe collected conversations with timestamps. Subsequently, Adobe Premiere Pro\footnote{\url{https://www.adobe.com/products/premiere.html}} was used to segment the corresponding video clips for alignment.

\paragraph{Data annotation} The annotation stage is of importance as it directly influences the dataset's utility, particularly in the context of supervised machine learning approaches. Given the ambiguous and diverse nature of sarcasm, many studies \cite{Castro,Gent,Chauhan_2020,Chauhan_2022,Ray, Zhang_CMMA} provide training or guidelines to annotators to ensure unbiased labeling. Some studies even enlist the assistance of professional linguists or experts for annotation. Conversely, a few studies \cite{Rakov,Burkhardt} opt for an unguided approach, avoiding any guidance to prevent potential biases and encourage subjective interpretations of sarcasm. Contextual information is often provided to annotators, considering the significance of context in sarcasm perception. Additionally, in the dataset development process, a minimum of two annotators is employed, and inter-annotator agreement (IAA) is calculated as a metric to assess data quality \cite{Tepperman,Bedi,Gent,Chauhan_2020,Chauhan_2022,Ray,Burkhardt, Zhang_CMMA}. Some researchers \cite{Rakov,Castro,Atassi_2010} also engage third parties, such as a third evaluator or survey participants, to evaluate the data's quality. During annotation, Kappa score \cite{Fleiss} is frequently used to measure the agreement level.

\begin{figure}[t]
    \centering
    \includegraphics[width=0.5\textwidth]{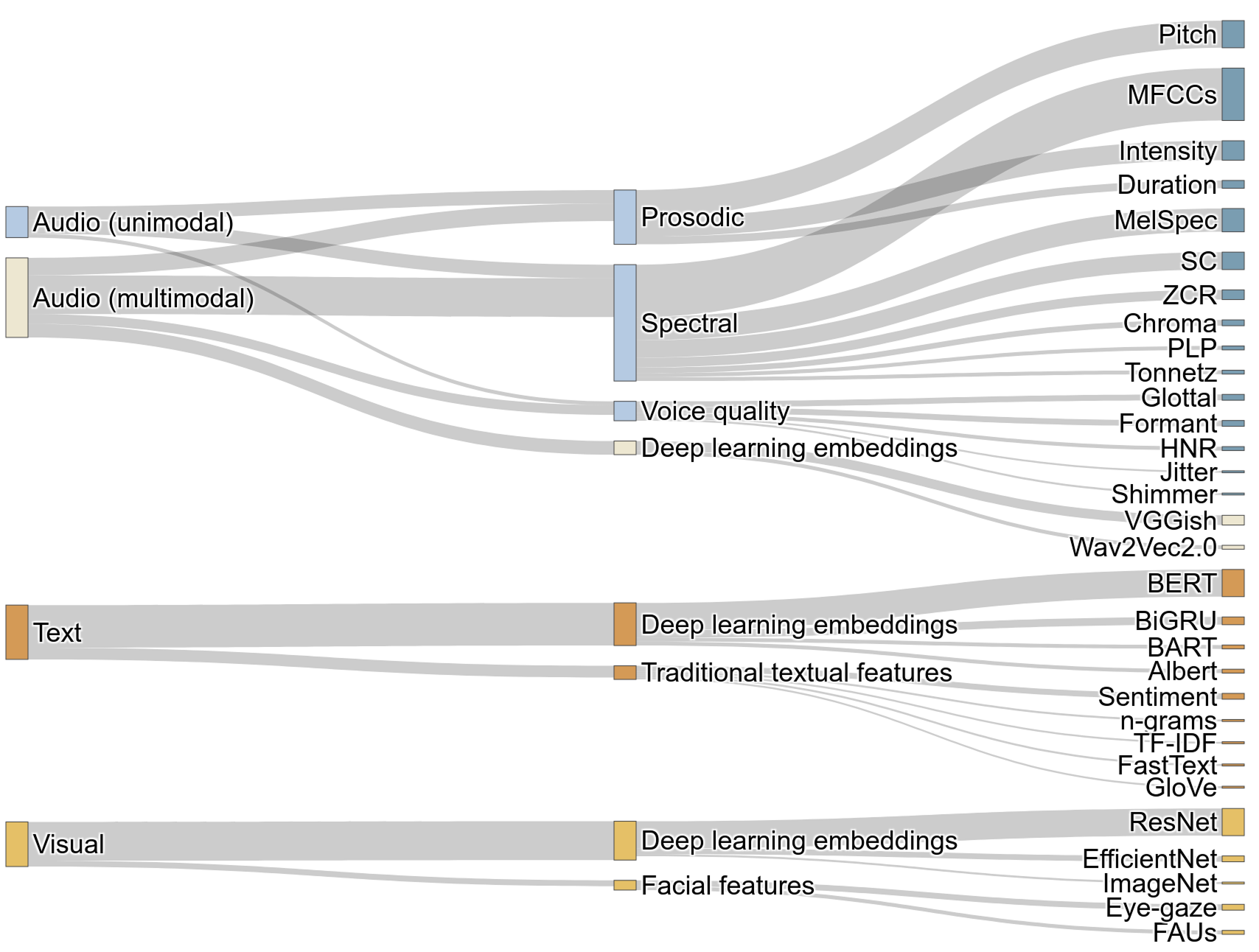}
    \caption{Hierarchical taxonomy of the most prominent features.}
    \label{feature_analysis}
\end{figure}

\subsection{Feature extraction analysis}
This section focuses on presenting insights pertinent to RQ2. To facilitate a more extensive understanding of the evolution from unimodal to multimodal sarcasm recognition, 
we have structured the relevant content into three categories: audio features, textual features and visual features. Each category allows us to delve into feature representations, the choice of toolkits used for feature extraction, and the feature extraction level that evolved from unimodal to multimodal contexts. Figure \ref{feature_analysis} summarizes key trends across unimodal and multimodal systems, presenting a hierarchical taxonomy of the most prominent features used in each modality. For a complete listing of specific features, classifiers, evaluation, and performance in the unimodal (audio-only) setting, see Appendix Table \ref{audio_features_unimodal}. For the corresponding summary in the multimodal setting, see Appendix Table~\ref{tablemultimodal}.

\subsubsection{Audio features}

We begin by examining the audio features employed in unimodal systems, as highlighted under Audio (unimodal) in Figure~\ref{feature_analysis}. We have classified the applied speech features into three distinct types as follows.

\paragraph{Spectral features} These features serve as a crucial representation of short-time speech signals, offering insights into the distribution of frequencies within a specific segment of the signal. These features are extracted through the application of Fast Fourier Transform, enabling the transformation of signal characteristics from the frequency domain into the time domain. In the context of sarcasm recognition, one of the prominent spectral features is Mel frequency cepstral coefficients (MFCCs). Six out of the eight unimodal studies reviewed have utilized this feature in their analysis. MFCCs are representations of the short-term power spectrum of a sound signal, and it is derived from the Mel frequency scale, which is a perceptually motivated frequency scale that better represents the way humans perceive the pitch of sounds. Therefore, they are well-suited for tasks involving human speech and sound recognition. More importantly, MFCCs offer a robust representation of spectral attributes within speech signals. For instance, Tepperman \textit{et al}. \cite{Tepperman} employed the first twelve MFCCs along with their delta and acceleration coefficients.

An alternative widely adopted choice is the Mel spectrogram (MelSpec), a time-frequency representation of audio employing the Mel frequency scale. Gao \textit{et al}. \cite{Gao_2022} utilized MelSpecs as input features for a convolutional neural network (CNN)-based architecture. This representation is highlighted for effective in capturing critical spectral nuances. Like MFCCs, MelSpecs are based on the Mel scale, making them suitable for speech and audio processing tasks. But they have higher dimensionality because of the retained spectral information in each frame. Furthermore, the application of Perceptual Linear Predictive (PLPs) features has been noted in Atassi \textit{et al}. \cite{Atassi_2010}, leveraging their specialization in mirroring the perceptual characteristics of speech as perceived by human listeners. In addition, a few studies \cite{Burkhardt, Sacheth} have extended their feature set to encompass additional spectral attributes, including chroma, Spectral Contrast (SCt), Spectral Rolloff (SR), Spectral Bandwidth (SBW), Tonnetz, and Zero Crossing Rate (ZCR). They demonstrated the effectiveness of applying spectral features in sarcasm recognition. 

\paragraph{Prosodic features} The features refer to the acoustic characteristics of speech related to its rhythm, melody, and intonation rather than its linguistic content. Koolagudi \textit{et al}. \cite{Koolagudi} divided prosodic features mainly in three categories: pitch, intensity and intonation. Rooted in variations in air pressure within the vocal fold, prosodic features serve as valuable indicators of human emotions \cite{Rao}. Many studies collectively applied statistical metrics such as mean, minimum, maximum, range, standard deviation to prosodic features, affirming their fundamental role in the context of sarcasm recognition in speech. For example, Atassi \textit{et al}. \cite{Atassi_2010} employed pitch contours, while Rakov and Rosenberg \cite{Rakov} incorporated word-level pitch and intensity contours, as well as sentence-level statistical measures of pitch, intensity, and speaking rate. Mathur \textit{et al}. \cite{Mathur} and Arun \textit{et al}. \cite{Arun} utilized a combination of spectral and prosodic features in their analyses.

\paragraph{Voice quality features} The features are recognized as the individual voice characteristic, including bandwidth, glottal attributes, harmonic-to-noise ratio (HNR), jitter, and shimmer. These features describe the manner in which an individual produces speech, relating to aspects of vocal cord vibration and the resultant acoustic output. To enhance feature representation, voice quality features are often combined with spectral and prosodic features. For example, in addition to the spectral and prosodic feature sets mentioned ealier, Atassi \textit{et al}. \cite{Atassi_2010} employed a voice quality feature set that included harmonicity, as well as the frequencies and bandwidths of the first three formants along with their first and second derivatives. Similarly, Burkhardt \textit{et al}. \cite{Burkhardt} utilized the larger-scale Interspeech 2013 Computational Paralinguistic Challenge (ComParE) feature set \cite{Schuller2013}, which encompasses a comprehensive range of features from the spectral, prosodic, and voice quality dimensions.

\paragraph{From unimodal to multimodal analysis}
In recent years, the multimodal approach has emerged as an effective solution for sarcasm recognition, offering a significant advancement over traditional methods that relied solely on text or audio. 
The audio features applied in these systems are summarized in Figure~\ref{feature_analysis}, under the category Audio (multimodal).

Notably, spectral features like MFCCs, MelSpec, ZCR, chroma, Spectral Centroid (SC), deltas and bandwidth have consistently gained prominence in the multimodal context \cite{Castro, Chauhan_2020, Alnajjar, Bedi, Ray, Hasan, Zhang_2021, Bharti, Ding, Sun, Pramanick, Wu, Pandey, Azahouani, Li_2024, Murthy, Manohar}. Among them, MFCCs are the most frequently applied features. The prosodic features have proven to play a vital complementary role as well. For instance, Alnajjar and Hämäläinen \cite{Alnajjar} and Hasan \textit{et al}. \cite{Hasan} applied pitches. Geng \textit{et al}. \cite{Geng} and Gent \textit{et al}. \cite{Gent} particularly utilized mean and average values of duration, pitch, intensity and speaking rate. Zhang \textit{et al}. \cite{Zhang_2021} collected pitch tracking, voice/unvoiced segment feature (VUVs), peak slope parameters, and maxima dispersion quotients (MDQ). Although less frequently employed, voice quality features have also been utilized in some studies. Hiremath and Patil \cite{Hiremath} utilized jitter, simmer, HNR; Hasan \textit{et al}. \cite{Hasan} included normalized amplitude quotient (NAQ), quasi open quotient (qOQ), glottal source parameters, harmonic model and phase distortions, and the formants in their speech feature set. Pramanick \textit{et al}. \cite{Pramanick} used glottal source parameters. Gent \textit{et al}. \cite{Gent} integrated HNR. 

A new trend in multimodal context is the use of deep learning models such as VGGish \cite{VGGish} and Wav2Vec2.0. \cite{wav2vec} VGGish is a CNN-based deep learning model, which takes MelSpec inputs and produces 128-dimensional embeddings that capture high-level semantic information from audio signals. Compared to traditional features (MFCCs, pitch and intensity), VGGish provides learned representations that encodes acoustic patterns without requiring manual feature design. However, it is trained primarily for general audio event classification, not for fine-grained speech nuances like sarcasm. Additionally, VGGish's fixed size embeddings might miss fine-grained temporal dynamics critical for detecting sarcasm. In contrast, Wav2Vec 2.0 is a transformer-based model trained using self-supervised learning on raw waveforms, producing contextualized frame-level embeddings that preserve both phonetic and prosodic information. 

Additionally, there has been a concerted effort to enhance feature representations using advanced techniques. A number of studies \cite{Hasan, Zhang_2023a} have explored the application of Transformer \cite{Vaswani}, and have shown promise in refining the extracted speech features and capturing intricate contextual information. Additionally, the integration of recurrent neural networks (RNNs) like the Gated Recurrent Unit (GRU) and attention mechanisms has been adopted in Zhang \textit{et al}. \cite{Zhang_2021}. Chauhan \textit{et al}. \cite{Chauhan_2022} took a similar approach, implementing Bidirectional GRU (BiGRU) \cite{Cho} to leverage contextual relationships and enhance feature representation. 

\begin{table}[h]
\centering
\caption{Toolkits for Speech Feature Extraction} \label{toolkitstable}
\footnotesize
\begin{tabularx}{0.48\textwidth}{
>{\raggedright\arraybackslash}p{1.8cm} 
>{\raggedright\arraybackslash}p{3.8cm} 
>{\raggedright\arraybackslash}p{2.2cm}} 
\toprule
\textbf{Toolkit} & \textbf{Features extracted} & \textbf{Coding language} \\
\midrule
\href{https://www.fon.hum.uva.nl/praat/}{Praat} & Spectral, pitch contour, formant contours, intensity contour, voice quality & Python, C++, Praat script \\
\href{https://audeering.github.io/opensmile/}{OpenSMILE} & Low-level and high-level spectral, temporal, intensity, pitch, voice quality & Python, C++ \\
\href{https://librosa.org/doc/latest/index.html}{Librosa} & Spectral, temporal, harmonics, intensity, pitch  & Python \\
\href{https://covarep.github.io/covarep/}{COVAREP} & Spectral envelop, pitch tracking, glottal features & MATLAB \\
\bottomrule
\end{tabularx}
\end{table}

\paragraph{Extraction toolkits}
Several toolkits have been developed for extracting speech features from audios; Table \ref{toolkitstable} lists the toolkits that are mentioned in the selected papers. The choice of an audio analysis toolkit depends on the specific sarcasm-related speech features being analyzed. Praat \cite{praat} excels in detailed and accurate phonetic analysis, offering features for pitch, formants, intensity, and spectral analysis, making it ideal for detecting tonal variations in sarcasm. openSMILE \cite{opensmile}, designed for real-time processing, efficiently extracts a broad range of emotion-related features, making it well-suited for sarcasm-related emotional cues like tone and intensity. 
Librosa \cite{librosa} integrates easily with Python and simplifies complex audio analyses, but it lacks sarcasm-related speech features such as speech rate. A recent addition to the toolkit inventory is COVAREP \cite{covarep}, which is an open-source toolkit that researchers can leverage to collect a rich set of features, including pitch, intensity, and spectral features. Each toolkit has its strengths and limitations, so choosing the appropriate tool should be guided by the specific sarcasm-related speech features being analyzed. 

\paragraph{Level of feature extraction}
Among the selected articles, two distinct categories of feature level emerge: local features and global features. Local features are characterized by their short-term nature, computed within small, often overlapping time windows in an audio signal. These features excel in capturing dynamic changes and nuanced variations in speech, as exemplified in the work of Tepperman \textit{et al}. \cite{Tepperman}. Conversely, global features are computed over more extensive segments of an audio signal, typically encompassing the entire speech utterance. These features capture characteristics that manifest over a more extended period, and a number of studies \cite{Atassi_2010,Arun,Mathur,Gao_2022} have employed global features in their experiments. It's noteworthy that a common practice is to initially extract local features from individual windows within an utterance and subsequently aggregate them across the entire utterance.
Nevertheless, it’s also applicable to combine these two types of features \cite{Rakov, Gent}. 

Our review reveals a notable inclination toward global features among researchers. This preference can be attributed to the comparative computational efficiency of global features, as they involve fewer parameters than their local counterparts. However, it's essential to acknowledge that extracting global-level features comes at the cost of disregarding word-level relationships within the utterance. From unimodal to multimodal settings, an intriguing development is the utilization of deep learning models to process local features, therefore achieve a global representation, effectively transforming the feature level. Bedi \textit{et al}. \cite{Bedi}, for instance, employed a 1D CNN to convert the speech features to a fixed-length of sequence. Zhang \textit{et al}. \cite{Zhang_2023a} used an average pooling layer and attention mechanism to generate the final feature vector. Ding \textit{et al}. \cite{Ding} applied a nonlinear mapping to achieve the same goal.

\subsubsection{Textual features}
It's seen from Figure \ref{feature_analysis} that a significant majority of research endeavors in multimodal sarcasm recognition have used word embeddings extracted from BERT \cite{Devlin} as a fundamental feature representation for the text modality \cite{Castro, Wu, Zhang_2021, Ding, Sun, Pramanick, Zhang_2023a, Liu, Zhang_2023b, Pandey, Zhang_CMMA, Gao_2024, Li_2024, Murthy}. BERT utilizes a bidirectional transformer architecture to pre-train deep bidirectional representations of text. Trained on extensive corpora like the Toronto Book Corpus \footnote{\url{https://github.com/sgraaf/Replicate-Toronto-BookCorpus}} and Wikipedia \footnote{\url{https://huggingface.co/datasets/legacy-datasets/wikipedia}}, BERT embeddings can capture nuanced semantic content. The incorporation of BERT embeddings in sarcasm recognition has gained prominence, as it equips models with the capacity to comprehend the intricate web of context and relationships among words. 
BART \cite{Lewis}, alongside its multilingual variation mBART are transformer architectures that combine the strengths of bidirectional and auto-regressive pre-training. This dual approach enables them to effectively grasp the contextual nuances and semantic meanings embedded in text, further enhancing the quality of extracted features.

Some studies have taken this approach a step further, enhancing word embeddings by integrating them with RNNs or transformer-based models. For instance, a few works \cite{Chauhan_2020, Chauhan_2022, Zhang_2023a, Azahouani} enriched the word embeddings with BiGRU. This addition allows them to to capture contextual information, enrich the semantics of text, and to generate a fixed-length of features. 
Furthermore, Liu \textit{et al}. \cite{Liu} introduced a complex-valued multi-modal encoder that leverages quantum probability to enrich features with contextual relationships, showcasing a novel approach to feature enhancement. 

Research in multimodal sarcasm recognition predominantly leans towards the use of global textual features. The preference for global features aligns with the trend observed in audio features. To transition from local to global features, researchers employ mathematical averaging or employ DNNs, facilitating the conversion of local feature information into a comprehensive global representation.

\subsubsection{Visual features}
As it's shown in Figure \ref{feature_analysis}, the visual modality predominantly relies on deep learning based spatial features \cite{Castro, Alnajjar, Ray, Wu, Zhang_2021, Ding, Pramanick, Zhang_2023a, Liu, Zhang_2023b, Pandey, Tomar, Zhang_CMMA, Li_2024, Tiwari}, which are essentially attributes of images capturing information related to the spatial distribution of pixels within each frame. These spatial features play a pivotal role in discerning the visual cues within the video content. With the help of DNNs such as ResNets \cite{resnet} and EfficientNet \cite{efficientnet}, the extracted features are assigned with semantic meanings, thus enriching their interpretability. 

Another category of extracted features in the visual modality pertains to facial features. Given the strong connection between sarcasm and facial expressions, using facial features directly avoids noisy features from the background or unrelated scenes of the video. For instance, Wu \textit{et al}. \cite{Wu} employed MTCNN \cite{MTCNN} for face detection in video frames. They then cropped these detected faces, subsequently processing them with a ResNet to extract features. Similarly, Sun \textit{et al}. \cite{Sun} utilized Face DLIB \footnote{\url{https://github.com/davisking/dlib}} for face extraction, followed by feature extraction using the ResNet-152. On the other hand, Hasen and Patil \cite{Hasan} implemented OpenFace 2 \cite{openface} to derive Facial Action Unit (FAU) features, which are crucial for analyzing human emotions and affect. Furthermore, Hiremath \textit{et al}. \cite{Hiremath} incorporated the eye aspect ratio metric in their analysis, adding another dimension to their facial feature domain. 

Several research endeavors have made contributions to feature representation by applying both preprocessing and post-processing techniques. For instance, Pramanick \textit{et al}. \cite{Pramanick} enhanced the spatial features by leveraging the action recognition tool 13D \cite{13d}. Meanwhile, Zhang \textit{et al}. \cite{Zhang_2023a} carried out further processing on spatial features, incoprprating BiGRU and the attention mechanisms, leading to the creation of weighted feature vectors as the final representation. Hasan \textit{et al}. \cite{Hasan} harnessed modified transformer encoder to enrich the facial features. 
To mitigate the inclusion of noisy features resulting from video frames, Ray \textit{et al}. \cite{Ray} implemented Katna \footnote{\url{https://katna.readthedocs.io/en/latest/}} on the extracted spatial features. This selective approach towards key frame extraction aids in enhancing the overall quality of the spatial feature. 

Similar to the audio and text modalities, global features are the preferred choice in the visual modality. Typically, these features are derived by averaging values across multiple video frames or images that are synchronized with the corresponding text and audio.
In a few instances, under the same mechanism, alternative methods have been employed. For example, Bedi \textit{et al}. \cite{Bedi} involved a 1D CNN for generating a global feature vector. Zhang \textit{et al}. \cite{Zhang_2023a} included average pooling and attention mechanisms to transform local features to global features. 

In addition to harnessing features from the audio, text, and visual modalities, Some studies \cite{Chauhan_2020, Chauhan_2022} ventured beyond the conventional boundaries by incorporating additional elements into their training sets. Specifically, they introduced speaker information, considering the identity of the speaker delivering the utterance as an influential factor in the recognition process. Chauhan \textit{et al}. \cite{Chauhan_2022} delved into this multidimensional approach by exploring the role of emojis in the context of sarcasm recognition, recognizing the potential significance of these visual cues in enhancing the overall performance of the task.

\subsection{Classification method and fusion}
The classification task typically involves using machine learning models to analyze the extracted features and determine whether it contains sarcasm. These classification models exhibit a spectrum of complexity, spanning traditional machine learning techniques such as Support Vector Machine (SVM) and Random Forest (RF) to the more advanced deep learning methodologies, which incorporate RNNs, CNNs, or Transformers. 

\begin{figure}[t]
    \centering
    \includegraphics[width=0.32\textwidth]{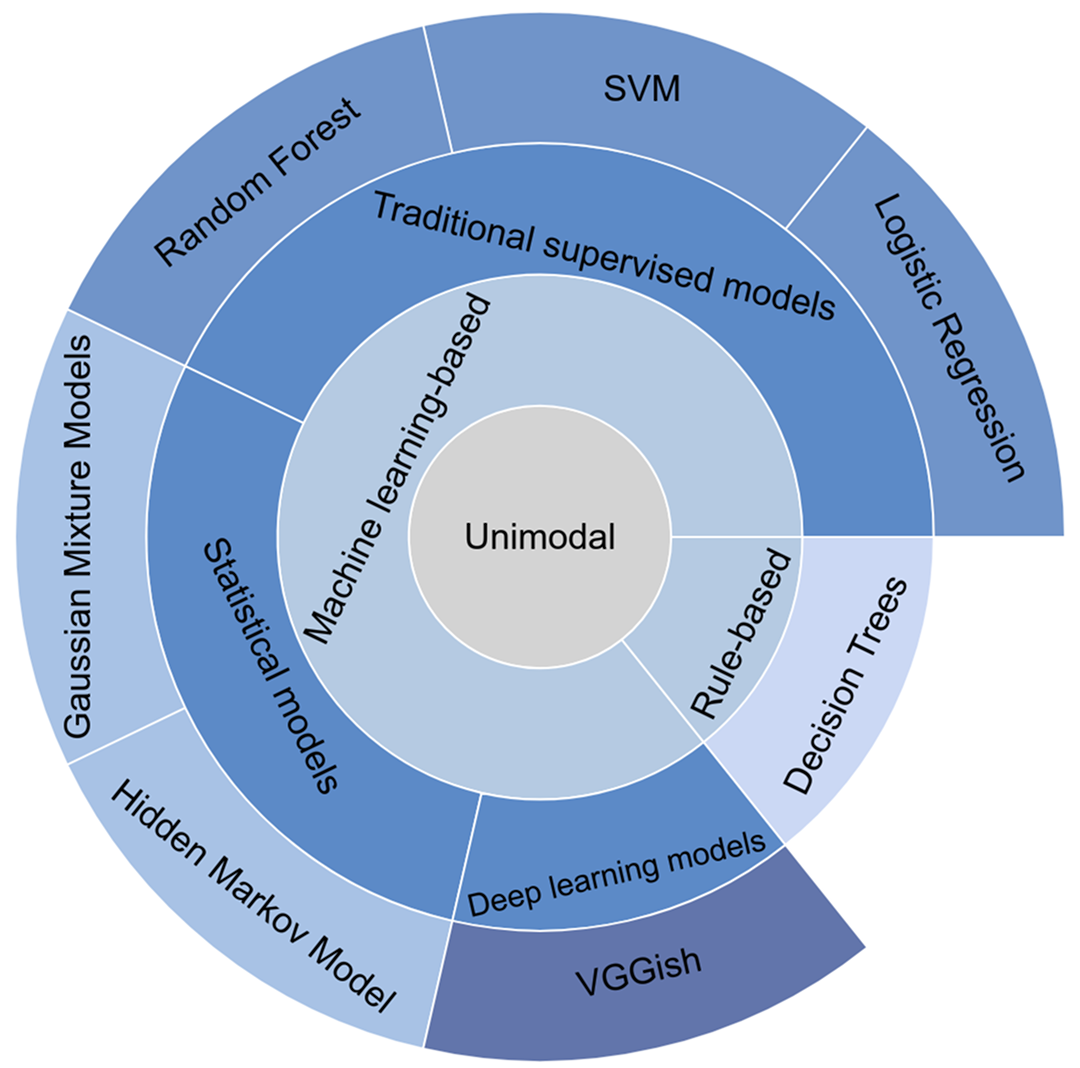}
    \caption{The overview of classification strategies in unimodal sarcasm recognition.}
    \label{classification_unimodal}
\end{figure}

\subsubsection{Unimodal classification}
Unimodal sarcasm recognition in the audio modality involves the application of various classifiers, each with its own unique characteristics. Figure \ref{classification_unimodal} provides a comprehensive overview. These utilized classifiers can be broadly categorized into two main groups:

\paragraph{Rule-based classifiers} Such classifiers as Decision Trees used by Tepperman \textit{et al}. \cite{Tepperman}, are known for their interpretability and rule-based structure. They can provide a clear insight into the decision-making process, making it a valuable tool for understanding the reasons behind classification outcomes. However, it may not handle complex data distributions effectively. 

\paragraph{Machine learning-based classifiers} 
Our review identified three main approaches to machine learning-based sarcasm classification.

\begin{itemize}
    \item \textbf{Statistical models} include classifiers like GMM proposed by Atassi \textit{et al}. \cite{Atassi_2010} and Hidden Markov Model (HMM) used by Mathur \textit{et al}. \cite{Mathur}. These models offer a principled and interpretable approach to capturing complex data distributions. GMMs are commonly used to model speech features because they can capture variations in human vocal expression. However, sarcastic speech may involve non-Gaussian patterns in feature space, and GMMs might not fully capture these nuances. On the other hand, HMMs are particularly effective for handling sequential data, making them well-suited for capturing temporal patterns in sarcasm. However, HMMs assume that the current state depends only on the previous state, which might not always capture more complex dependencies.
    \item  \textbf{Traditional supervised learning models} encompass a spectrum of classifiers, such as Logistic Regression (LR) \cite{Rakov}, SVM \cite{Bharti}, RF \cite{Arun}. These classifiers leverage established techniques within the domain of supervised learning to make informed decisions. LR is valued for its simplicity and interpretability, making it particularly effective for binary classification problems. However, it performs poorly when dealing with complex, non-linear relationships. SVM is known for the ability to find optimal hyperplanes that separate different classes, even in high-dimensional spaces. RF, an ensemble learning method, combines multiple decision trees to improve accuracy and robustness. It can provide valuable insights into the importance of different features in making predictions. 
    \item  \textbf{Deep learning models} encompass advanced models such as the fine-tuned VGGish employed in Gao \textit{et al}. \cite{Gao_2022}. These models harness CNN to capture intricate patterns in the audio data. Due to their deep architecture, they can model intricate relationships within the data. However, training deep learning models requires a substantial amount of labeled data. Acquiring and annotating large datasets can be resource-intensive and time-consuming. In addition, DNNs are often considered ``black boxes'' due to their complex architectures, making it difficult to interpret how they arrive at specific decisions or predictions. 
\end{itemize} 

\subsubsection{Multimodal fusion strategies}
Recent experimental studies have consistently demonstrated that incorporating audio, textual, and visual data significantly enhances sarcasm recognition performance compared to unimodal approaches, highlighting the inherently multimodal nature of sarcastic expressions \cite{Castro, Hasan, Wu, Ray}. This reflects the linguistic understanding that sarcasm often relies on the interplay of audio, textual and visual cues – an interaction that is difficult to capture through a single modality alone \cite{Attardo, Hancock}. 
As pointed by Jacob \textit{et al.} \cite{Jacob_2016}, the incongruity between verbal and non-verbal cues is facilitating comprehension of sarcasm. Sarcasm is neither simply a tone of voice or verbal irony, its complexity emerges from the dynamic integration of multiple communication channels. 

Furthermore, the multimodal nature of sarcasm is supported by the taxonomy provided by Attardo \cite{Attardo}, which categorizes sarcasm markers into two primary types. \textit{Metacommunicative alerts} are explicit signals that directly inform the listener that an utterance should be interpreted sarcastically. These include verbal indicators such as ``just kidding'' or ``I'm being sarcastic,'' as well as non-verbal cues such as a sarcastic smile or tongue-in-cheek expression. These markers serve as overt cues, providing a clear signal of the speaker's sarcastic intent. In contrast, \textit{paracommunicative alerts} are more subtle, involving indirect cues that, when paired with the literal statement, suggest a sarcastic interpretation. For example, a speaker might use a blank facial expression or exaggerated nodding to imply sarcasm, while phonetic cues could include delivering an enthusiastic statement in a bored or monotonous tone or using flat intonation with minimal pitch variation. Unlike metacommunicative signals, paracommunicative alerts create a contrast with the literal content, prompting the listener to infer sarcasm.
\begin{figure}[t]
    \centering
    \includegraphics[width=0.43\textwidth]{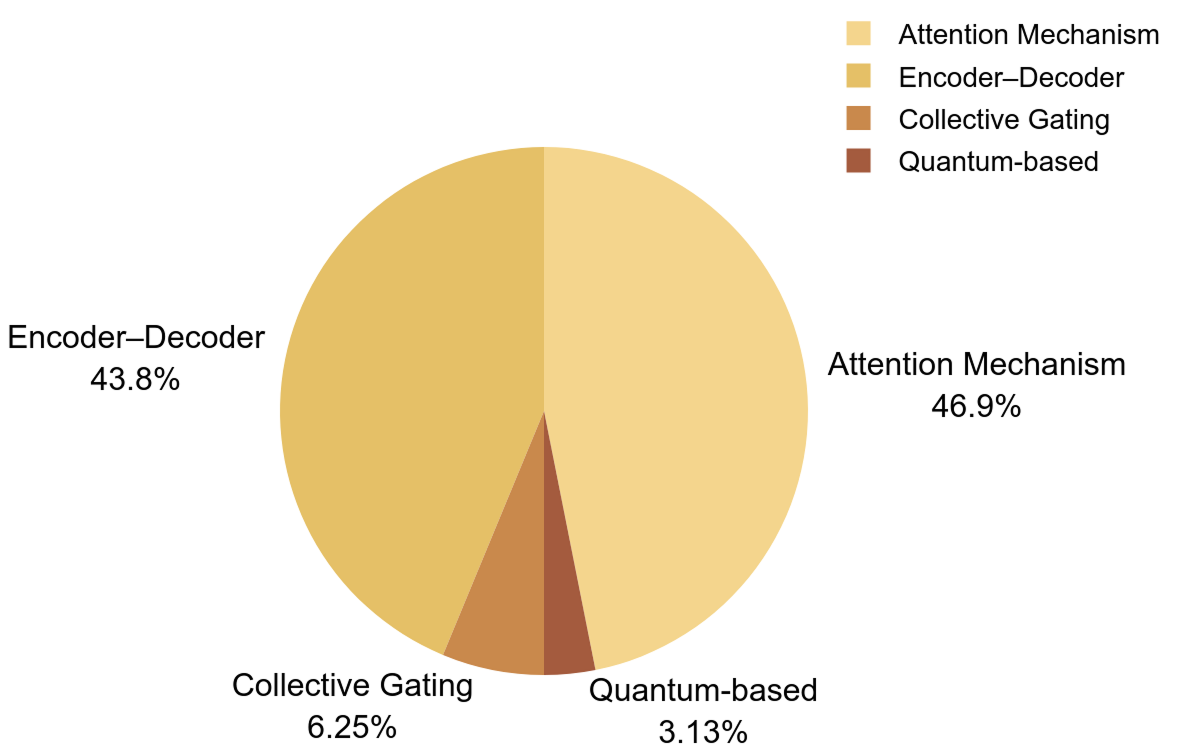}
    \caption{The distribution of current multimodal fusion strategies.}
    \label{fusion}
\end{figure}
Furthermore, different modalities contribute varying degrees of significance to sarcasm detection. For instance, some speakers rely heavily on prosodic features in the audio modality, such as pitch variations, to convey sarcasm, while others may primarily use semantic cues embedded in the text modality. This underscores the value of multimodal integration in resolving ambiguity and improving robustness in sarcasm recognition systems.

The effectiveness of multimodal sarcasm recognition hinges on appropriately weighting these different modalities to reflect their respective contributions. To present this in detail, we delve into the various model fusion techniques in the field, discussing their roles in enhancing the effectiveness of multimodal sarcasm recognition systems. Figure \ref{fusion} presents the distribution of the fusion strategies. For a complete listing of fusion methods, see Appendix Table \ref{tablemultimodal}.

Traditional taxonomies classify fusion methods into three categories: \textit{Early fusion}, where raw data from each modality is combined before being input to the model; \textit{Intermediate fusion}, where features extracted from various modalities are merged and then fed into the model for decision-making; \textit{Late fusion}, where decisions made independently by each modality are combined to produce the final prediction. 
However, traditional classifications do not adequately capture the complexities of contemporary architectures, where feature representation, modality integration, and classification are intricately intertwined. To address these complexities, we adopt the more recent taxonomy proposed by Zhao \textit{et al}. \cite{Zhao}, which provides a more nuanced framework for understanding the fusion techniques employed in multimodal sarcasm recognition.

\paragraph{Encoder-decoder}
This architecture is structured into two primary components: the encoder and the decoder. The encoder processes input data into a latent representation, retaining the essential semantic information while filtering out extraneous noise. The decoder then utilizes this latent representation to generate predictions. In our review, a number of studies have adopted this approach. For instance, Castro \textit{et al}. \cite{Castro} used encoders to extract textual, speech, and visual representations, which were then concatenated and fed into a SVM decoder for classification. 

Other variants within this framework include Ding \textit{et al}. \cite{Ding}, who proposed a multimodal learning framework consisting of three layers of feature fusion networks for encoding text, audio, and visual modalities. The output of the third layer was then combined with speaker features and contextual information before being passed to a fully connected neural network (FCNN) decoder for final prediction. Gent \textit{et al}. \cite{Gent} utilized a Long Short-Term Memory (LSTM) network to encode audio features, and a CNN to encode the textual data, with the resultant representations concatenated and processed by a FCNN functioning as the decoder. 

\paragraph{Attention mechanism}
Introduced by Vaswani \textit{et al}. \cite{Vaswani}, the attention mechanism has become a central topic in the deep learning community. This mechanism allows models to assign varying weights to different modalities, enabling the extraction of information that is critical to the task at hand. By doing so, attention mechanisms enhance prediction accuracy without significantly consuming computational costs. Recently, attention mechanisms have emerged as key tools in multimodal data fusion tasks. Several variations of attention-based methods have been developed to leverage this capability. 
    
First, the \textit{intra-modality self-attention} focus exclusively on data within a single modality, facilitating a undiluted analysis of relationships within that modality. Pramanick \textit{et al}. \cite{Pramanick} proposed the MuLOT system, which utilizes self-attention to enhance intra-modal correspondence while incorporating optimal transport methods to address cross-modal alignment.
    
Second, the \textit{inter-modality cross-attention} focuses on exploiting the relationship among different modalities by integrating data from multiple sources in each attention operation. This approach enables the model to capture intricate connections across modalities. For example, Chauhan \textit{et al}. \cite{Chauhan_2020} employed cross-attention to discern relationships between segments across modalities. Hasan \textit{et al}. \cite{Hasan} and Zhang \textit{et al}. \cite{Zhang_2023a} utilized a pair of cross-modal attention mechanisms to explore bimodal interactions (text-audio, text-visual interactions). Wu \textit{et al}. \cite{Wu} applied a cross-modal attention mechanism to emphasize specific words, particularly those exhibiting a high degree of incongruity between positive spoken text and negative nonverbal (audio and visual) cues. Similarly, Zhang \textit{et al}. \cite{Zhang_2021} proposed a framework that assessed cross-modal incongruity between text and visual, as well as text and audio modalities. Furthermore, Chauhan \textit{et al}. \cite{Chauhan_2022} developed an architecture to explore the interplay between emojis and other multimodal data, highlighting the versatility and evolving nature of cross-attention techniques.
    
Third, the \textit{hybrid intra-modality and inter-modality attention} combines the previous two. 
Zhang \textit{et al}. \cite{Zhang_2023a} implemented an intra-modality attention mechanism to identify the most relevant features within each modality individually. Following this, they employed three inter-modal attention mechanisms to capture and analyze the pairwise relationships between modalities (text-visual, text-audio, and visual-audio interactions). This dual-layered approach enables a comprehensive understanding of how different types of data contribute to sarcasm detection, facilitating a more nuanced and effective integration of multimodal information. 
    
  
\paragraph{Collaborative gating mechanism}
Unlike attention mechanism which distributes weights across elements within or across modalities, gating mechanisms leverage learned gates to control which modality influences the fused representation. For example, Ray \textit{et al.} \cite{Ray} and Tomar \textit{et al.} \cite{Tomar} used the collaborative gating mechanism to evaluate all modalities in parallel and learned to assign importance scores that reflect their combined informative strength. This coordinated gating strategy enables the model to selectively attenuate less relevant or noisy modalities while amplifying those contribute saliently.
    
\paragraph{Quantum based method}
Last, beyond the multimodal fusion methods previously discussed, innovative approaches are emerging to enhance multimodal fusion. For instance, Liu \textit{et al}. \cite{Liu} introduced QUIET, an architecture featuring a \textit{quantum-based} inter-modal fusion layer designed to integrate multimodal data. At the heart of this framework, bimodal representations are derived through quantum interference, a process grounded in quantum probability theory. Quantum probability offers a mathematical and conceptual foundation for modeling the inherently uncertain behaviors of microscopic particles, providing a novel mechanism for capturing complex inter-modal relationships. Tiwari \textit{et al.} \cite{Tiwari} proposed QFNN that incorporates quantum computation to enhance feature expressiveness. Furthermore, quantum entanglement is utilized to model interactions between modalities, capturing nuanced dependencies that are difficult to express in classical space.
    
To summarize, the strength of encoder-decoder methods lies in their ability to integrate and aggregate features from multiple abstraction levels, thereby enhancing the model's capacity to capture cross-modal relationships. However, multiple encodes for different modalities demand substantial computational resources, especially when the number of modalities is increased. Attention mechanisms play a crucial role in facilitating interactions both within and across modalities, refining features by leveraging learned associations. This process enhances the model’s ability to capture and understand the underlying relationships within individual modalities and the intricate connections between them. However, this approach inherently introduces structural complexity, as it requires the seamless integration of diverse data types, each with its own unique characteristics. Ensuring that these modalities are high-quality and properly synchronized is critical to optimizing the attention process and accurately representing the nuanced relationships between them. Moreover, gating mechanisms act as dynamic filters, allowing the model to selectively focus on the most relevant features from each modality while limiting less informative ones. Furthermore, by leveraging quantum computing principles, such as superposition and entanglement, quantum based methods can potentially handle complex, high-dimensional multimodal data.

\subsubsection{Performance meta-analysis}


We conducted a meta-analysis to aggregate F1-scores from multiple studies on sarcasm recognition, offering a comprehensive view of fusion method performance. Results are reported separately for speaker-dependent and speaker-independent settings, as each presents unique challenges: speaker-dependent involves the same speakers for training and testing, while speaker-independent employs different speakers, providing a more realistic test of model generalizability. The following sections detail the data extraction and analysis process.

\paragraph{Data extraction}
To assess the effectiveness of different fusion methods, we first selected studies with consistent contexts, such as dataset used, reported metrics, and evaluation methods. Initially, 31 multimodal sarcasm recognition studies were gathered. We refined the selection using inclusion and exclusion criteria: Studies employing the MUStARD dataset or its extensions were retained (22 articles). Given sarcasm recognition involves class imbalance, studies reporting F1-score, a balanced metric, were included (19 articles).  We excluded five studies without explicit evaluation methods, and five used train-test splits that were insufficient for stable statistical analysis, resulting in nine articles for the analysis. 

\paragraph{Analysis}
The primary goal is to determine whether major fusion methods (encoder-decoder VS. attention mechanism) significantly impact F1-scores in sarcasm recognition. To achieve this, we conducted a random-effects meta-analysis, assessing the heterogeneity across studies and computing effect sizes to quantify the magnitude of differences in F1-scores between the fusion methods. The effect size was calculated using $Hedges' g$ \cite{Hedges}, which adjusts for small sample sizes and quantifies differences in mean F1-scores. We used the $Q$-statistic and $I^2$ statistic to assess heterogeneity across studies. The $Q$-statistic tests whether the variation in effect sizes across studies is greater than expected by chance, while $I^2$ quantifies the percentage of total variation due to heterogeneity \cite{Higgins}. After assessing heterogeneity across studies, a random-effects model was applied. The model incorporated tau-squared ($\tau^2$), an estimate of the between-study variance, and provided a pooled mean effect size \cite{DerSimonian}.

\begin{table}[t]
\centering
\caption{Performance of speaker-independent 5-fold cross-validation} \label{table_eva_a}
\footnotesize
\begin{tabularx}{0.45\textwidth}{
>{\raggedright\arraybackslash}p{3.0cm} 
>{\raggedright\arraybackslash}p{1.8cm} 
>{\raggedright\arraybackslash}p{4cm}} 
\toprule
\textbf{Fusion method} & \textbf{F1-score (\%)} & \textbf{Reference} \\
\midrule
Encoder-decoder  & 63.1 &  SVM decoder \cite{Castro} \\
Encoder-decoder  & 71.3 & FCNN decoder \cite{Ding} \\
Attention mechanism  & 65.90 & cross-attention \cite{Chauhan_2020} \\
Attention mechanism & 70.00 & cross-attention \cite{Wu} \\
Attention mechanism  & 70.90 & cross-attention \cite{Li_2024} \\
Attention mechanism  & 80.00 & cross-attention \cite{Chauhan_2022} \\
\bottomrule
\end{tabularx}
\end{table}

Table \ref{table_eva_a} summarizes studies with speaker-independent evaluation. We applied the aforementioned analysis framework to the speaker-independent evaluation and observed significant heterogeneity across studies, with a $Q$-statistic of $30.00$ and an $I^2$ of $83.33\%$. Using the random-effects model, the weighted mean F1-score for the encoder-decoder method was $67.20$, with a 95\% confidence interval of $(59.16, 75.24)$. In comparison, the attention mechanism yielded a higher mean F1-score of $71.70$, with a 95\% confidence interval of $(65.87, 77.53)$. However, the overall $p$-value for the difference between the two methods was $0.4630$, suggesting that the difference in performance was not statistically significant.

\begin{table}[t]
\centering
\caption{Performance of speaker-dependent 5-fold cross-validation} \label{table_eva_b} 
\footnotesize
\begin{tabularx}{0.45\textwidth}{
>{\raggedright\arraybackslash}p{3.0cm} 
>{\raggedright\arraybackslash}p{1.8cm} 
>{\raggedright\arraybackslash}p{4cm}} 
\toprule
\textbf{Fusion method} & \textbf{F1-score (\%)} & \textbf{Reference} \\
\midrule
Encoder-decoder & 70.35 & SVM decoder \cite{Bharti} \\
Encoder-decoder & 71.60 & SVM decoder \cite{Castro} \\
Encoder-decoder  & 75.01 & FCNN decoder \cite{Ding} \\
Encoder-decoder  & 79.00 & FCNN decoder \cite{Sun} \\
Attention mechanism  & 72.57 & cross-attention \cite{Chauhan_2020} \\
Attention mechanism  & 74.50 & cross-attention \cite{Wu} \\
Attention mechanism  & 78.70 & cross-attention \cite{Chauhan_2022} \\
Attention mechanism  & 74.60 & hybrid attention \cite{Zhang_2023a} \\
Attention mechanism  & 75.20 & hybrid attention \cite{Li_2024} \\
\bottomrule
\end{tabularx}
\end{table}

Table \ref{table_eva_b} covers those with speaker-dependent evaluation. The speaker-dependent evaluation also demonstrated substantial heterogeneity, with a $Q$-statistic of $72.00$ and an $I^2$ of $88.89\%$. Using the random-effects model, the weighted mean F1-score for the encoder-decoder method was $73.99$, with a 95\% confidence interval of $(70.19, 77.79)$. In comparison, the attention mechanism achieved a comparable mean F1-score of $75.11$, with a 95\% confidence interval of $(73.15, 77.07)$. The overall $p$-value for the difference between the two methods was $0.6303$, confirming no statistically significant difference.

The analysis shows no statistically significant difference in performance between attention mechanism and encoder-decoder fusion methods in both speaker-independent and speaker-dependent evaluations. While attention mechanism exhibits higher mean F1-scores in both cases, the overlapping confidence intervals and high $p$-values indicate that these differences are not statistically reliable. Significant heterogeneity among studies suggests that model performance can be affected by other factors like feature granularity and model architecture. Additionally, the limited study pool restricts statistical power, highlighting the need for more research to detect subtle differences between fusion methods.
\begin{table}[t]
\centering
\caption{Examples of common failure cases} \label{table_examples} 
\footnotesize
\begin{tabularx}{0.49\textwidth}{
>{\raggedright\arraybackslash}p{4.4cm} 
>{\raggedright\arraybackslash}p{0.9cm} 
>{\raggedright\arraybackslash}p{0.9cm}
>{\raggedright\arraybackslash}X
} 

\toprule
\textbf{Utterances} & \textbf{Actual} & \textbf{Predicted} & \textbf{Reference} \\
\midrule
1. Oh yeah, ok, including the waffles last week, you now owe me seventeen zillion. & Sarcasm & Non-sarcasm & \cite{Chauhan_2020} \\
2. Do cocaine smugglers write ‘cocaine’ on the box?  & Sarcasm & Non-sarcasm &  \cite{Ding} \\
3. Yes, you can. You’re thinking about time, you can’t go back in time.  & Non-sarcasm & Sarcasm & \cite{Chauhan_2022} \\
4. I thought if I littered, that crying Indian might come by and save us.  & Sarcasm & Non-sarcasm & \cite{Chauhan_2022} \\
\bottomrule
\end{tabularx}
\end{table}
\paragraph{Error analysis and failure modes}
While aggregate F1-scores offer insights, understanding misclassification reasons is essential for refining model performance. Below, we examine common failure modes in sarcasm recognition:

\begin{itemize}
    \item Mismatch between modalities: Sarcasm is misclassified due to mismatches between modalities. For instance, Castro \textit{et a} \cite{Castro} noted errors arising when facial expressions and emotional cues conflicted. Similarly, Chauhan \textit{et al.} \cite{Chauhan_2020} observed confusion between explicit and implicit emotions, as in utterance 1 in Table \ref{table_examples}, where Chandler’s happy demeanor (explicit emotion) contrasts with his sarcastic anger (implicit emotion), leading to misclassification.
    \item Neutral expressions and monotone speech: Sarcasm is challenging to detect when visual and vocal cues are neutral. Ding \textit{et al.} \cite{Ding} highlighted Sheldon's expressionless, monotone speech in sentence 2 (Table \ref{table_examples}) as difficult for models to interpret without added context. Chauhan \textit{et al.} \cite{Chauhan_2022} also found models failed with neutral sentiment and expressionless emojis, as these cases rely heavily on contextual or commonsense knowledge that models often lack.
    \item Nuanced sarcastic expressions: Labeling limitations can obscure subtle sarcasm. Chauhan \textit{et al.} \cite{Chauhan_2022} noted cases where sentences with identical sentiment and emoji annotations were classified differently. For example, utterance 3 (Table \ref{table_examples}) was labeled as non-sarcastic while sentence 4 with the same annotations was sarcastic, leading to model confusion. Limited training data further hinders the ability to capture these nuanced cues.
\end{itemize}

To summarize, key challenges identified include modality mismatches, such as inconsistencies between facial expressions and expressed emotions, and difficulties detecting sarcasm with neutral cues. Additionally, models struggle when labeling fails to capture sarcasm's subtlety, indicating a need for more sophisticated multimodal fusion techniques. Existing labeling systems may not fully encapsulate sarcasm's complexity; developing more granular systems that incorporate factors like sarcasm intensity and confidence level, beyond sentiment and emotion, could help reduce misclassification.

\section{Discussion}
In this discussion, we synthesize key findings from our review to highlight emerging trends in sarcasm recognition research. Specifically, we identify evolving patterns in data curation, advances in feature extraction techniques, and the increasing sophistication of multimodal fusion architectures. We also examine limitations in each of these dimensions, based on which, we outline promising future directions, arguing that these trends and constraints indicate the need for more interdisciplinary and linguistics informed approaches. 

\subsection{Challenges in existing datasets}
\noindent \textbf{RQ1.1:} \textit{What are the available datasets for sarcasm recognition using speech data, and what limitations do they present?}

In our study, we have collected a total of nine primary datasets for the purpose of sarcasm recognition across different languages. These datasets include MUStARD, IITKGP-SEHSC, SEmoji MUStARD, The Best Sarcasm Annotated Dataset in Spanish, SEEmoji MUStARD, Irony-Recognition, MUStARD++, MaSaC and CMMA. These datasets empower multimodal research, augmenting the effectiveness of sarcasm recognition. Additionally, they can be used for multitask research due to extensive sentiment and emotion labelling. 
However, several areas for improvement have been identified in our review: (a). None of the existing datasets is derived from spontaneous speech - naturally occurring speech produced in real-life contexts without preparation or scripting. This type of data is critical for training sarcasm recognition systems that can function effectively in real-world applications. (b). The datasets available for sarcasm recognition are relatively small compared to those in related fields such as SA or Speech Emotion Recognition (SER), which may limit the generalizability and robustness of the models trained on them. (c). The lack of language diversity restricts the scope of multilingual sarcasm research, potentially overlooking the cultural nuances inherent in sarcastic communication. Future work should prioritize the development of spontaneous, multimodal, richly labeled, and culturally diverse datasets that are accessible for global research efforts.
(d). To establish the golden standard, some previous works employed participants who were not given any definition of sarcasm, leaving subjectivity in the generated dataset. The annotation generated by Tepperman \textit{et al}. \cite{Tepperman} received an IAA of 0.527 which is just above the chance. Applying such a dataset to evaluation is doubtful. (e). COST 2102 Italian Database of Emotional Speech \cite{Esposito} is a well-curated dataset comprising 216 samples labeled for various emotional states, including happiness, sarcasm/irony, fear, anger, surprise, and sadness. Moreover, Geng \textit{et al}. \cite{Geng} compiled a dataset containing Mandarin acted speech for sarcasm recognition. Regrettably, these resources are not available for open-source use or public access, limiting their utility in broader research contexts.\\

\noindent \textbf{RQ1.2:} \textit{What guidelines should be followed for creating high-quality datasets in this field?} \\
Our investigation into dataset development underscored the significance of data sources and the annotation process. 

\subsubsection{Data source} Our review highlights a growing trend of utilizing TV series as a source of data. Such data source achieves a good balance between collection efficiency and data spontaneity. However, the performance may be compromised by a lack of authenticity, and the collected data frequently encounters challenges such as background noise and laugh tracks. Several notable issues have been identified with the widely used MUStARD dataset. For example, the presence of background laugh tracks often overlaps with sarcastic speech, introducing potential bias into models trained on this data. Moreover, many utterances in the dataset are less than one second long and consist of only a single syllable (e.g., ``No''), making them dependent on preceding conversational turns for proper interpretation. Training models on such isolated utterances without the context of prior dialogue may lead to biased results. Furthermore, the character Sheldon, who is a major contributor of sarcastic remarks in the dataset, is portrayed as someone who struggles to recognize sarcasm. As a result, his sarcastic remarks are often minimally marked, affecting the presence of non-verbal cues such as prosody and facial expressions, which can further skew the dataset’s representation of sarcasm. The source of the dataset is of significant importance as it reflects nuanced facets of sarcasm. Diverse sources, such as social media dialogues, public debates could potentially offer a broader spectrum of sarcasm nuances. 

\subsubsection{Data annotation} It becomes evident from our review that providing annotators with guidelines for labeling sarcasm is crucial to ensure the consistency and accuracy of the annotation process. According to Camp \cite{Camp}, sarcasm encompasses a spectrum of sub-categories, each with its distinctive nuances, including \textit{Propositional Sarcasm}, which is an implicit sentiment proposition, relying on contextual information for sarcasm identification. For instance, ``your plan is fantastic!'' may sound non-sarcastic if the context is missing.  \textit{Illocutionary Sarcasm}, which mainly relies on non-verbal cues, such as prosody and visual signals, to convey sarcastic intent instead of the textual information alone. An example of this would be saying ``that’s right'' while rolling one's eyes or placing exaggerated emphasis on the word ``right.'' \textit{Embedded Sarcasm}, which is characterized by the presence of embedded sentiment incongruity; and \textit{Like-prefixed Sarcasm}, which introduces an implicit disagreement and it typically involves the use of the word ``like'' as a precursor to the sarcastic expression. 
Establishing annotation guidelines that encompass these nuanced sub-types of sarcasm is essential for creating well-organized and high-quality datasets. Moreover, this detailed classification can inspire the development of models that capture the intricacies of sarcasm and provide a more comprehensive analysis of model performance. 

Considering the cultural variability in interpreting sarcasm, it is imperative to provide annotators with comprehensive guidelines. These guidelines should not constrain interpretations but offer a broad framework for understanding sarcasm across diverse cultural and linguistic contexts. Currently, the IAA of the existing datasets is still at a low point, with Ray \textit{et al}. \cite{Ray} achieving the existing highest score of 0.595 in English. Therefore, it’s necessary to allow annotators to engage in discussions to establish a common annotation standard and increase the agreement. 

The current approach to sarcasm labeling is limited to a binary classification, where utterances are categorized as either sarcastic or non-sarcastic. This simplistic labeling neglects the subtleties and gradations of sarcasm that occur in real-world communication. In reality, sarcasm exists along a spectrum, with varying degrees of intensity. For example, when someone arrives late to a meeting, a remark like ``Oh, it’s not like we were waiting for you,'' delivered in a calm, flat tone, subtly conveys mild annoyance or disappointment through sarcasm. In contrast, responding to someone failing a simple test with, ``Congratulations! You’ve really outdone yourself this time!'' in an exaggeratedly excited tone amplifies the sarcasm. Sarcasm, often linked to negative sentiment, intensifies as the expression becomes more positive in valence and higher in arousal. In other words, the stronger the expressed emotion, the greater the perceived intensity of the sarcasm. Although the MUStARD++ dataset includes emotion annotations, it does not explicitly indicate the relationship between sarcasm and the annotated emotions. Therefore, to more accurately reflect real-life sarcasm, a labeling system that accounts for varying intensities is essential. This nuanced approach would better capture the range of sarcastic expressions, providing a more comprehensive framework for sarcasm analysis. 

\subsection{Evolution of feature extraction techniques}
\noindent \textbf{RQ2:} \textit{As the field progresses from unimodal to multimodal sarcasm recognition, how have feature extraction developed? What are the challenges and limitations in current feature extraction methods?}

In our exploration of the progression from unimodal to multimodal approaches for sarcasm recognition, 
we find three main trends: the increasing reliance on deep learning-based features, pre- and post-extraction refinement techniques, and the prevalent use of global feature representations. 

\subsubsection{Deep learning-based features}
In the transition to multimodal sarcasm recognition, the audio modality continues to play a pivotal role, but a shift has emerged. Recent trends indicate a growing adoption of DNNs to generate higher-level, semantically-rich speech features for audio feature extraction. We discover that several recent works have applied VGGish and Wav2Vec2.0, which allow for the representation of semantic meanings, marking a departure from earlier approaches that predominantly relied on tools like Librosa, OpenSMILE or Praat to extract low-level features. 

This shift towards higher-level feature representation is even more pronounced in the text modality, where sarcasm detection has been a long-standing area of interest. By comparing the feature extraction (e.g., syntactic pattern, n-grams, sentiment-related features, etc.) mentioned in the review conducted by Joshi \textit{et al}. \cite{Joshi} and the contemporary techniques utilized for text feature extraction (e.g., BERT, BiGRU, etc.), we gain insights into the potential trajectory of feature extraction within the audio modality. This progression suggests a future where feature extraction in the audio domain may follow a similar trajectory, moving towards the integration of more advanced and semantics embedded methods. 

Visual features were integrated as the rise of multimedia content on social media. 
Our review underscores the prevalence of spatial visual features. These features are systematically extracted through the utilization of DNNs, including architectures like ResNets and EfficientNet. The choice of these DNNs is primarily motivated by their robustness and exceptional performance in feature extraction. Importantly, it is evident from our investigation that pre-trained models, specifically from CNNs pretrained on extensive image datasets, has emerged as a prevailing practice in video feature extraction. 

\subsubsection{Pre- and post-extraction enhancements}
In addition to the increasing application of DNN-based features, many studies have employed various techniques to improve feature representation. In both audio and text modalities, transformer-based modules or RNNs are integrated into the architecture after feature extraction to enhance the ability to capture contextual relationships, enrich semantic content, and generate fixed-length feature representations. For the visual modality, beyond the use of transformer-based post-extraction modules, several pre-extraction techniques are applied directly to raw data to strengthen visual representation. These include methods such as action recognition and face recognition. 

\subsubsection{Prevalence of global features}
Our analysis regarding feature level, whether local or global, distinctly reveals a strong inclination among researchers towards the utilization of global features. In contrast to local features, which are derived from each individual segment and thus remain stationary, global features are statistical aggregates of these local components. Their prevalence in the literature is primarily attributed to their computational efficiency. However, it’s worth noting that global features have certain limitations. They overlook the word-level dynamics and temporal intricacies within an utterance, which can be crucial for capturing cues such as word stress, particularly in the context of sarcasm. Fortunately, in the context of the multimodal approach, our review uncovers a trend where researchers have introduced innovative techniques to enhance the extraction of global features. Rather than resorting to the conventional method of averaging feature vectors, these studies have opted for alternative approaches. For instance, the employment of 1D CNN to convert the speech feature \cite{Bedi}, or the utilization of attention mechanism to generate the feature vector \cite{Zhang_2023a}. These novel methods represent a valuable departure from the conventional reliance on averaged global features, thereby opening up new avenues for more effective feature extraction while accommodating the intricate temporal aspects involved in sarcasm recognition.

\subsubsection{Limitations of current feature extraction techniques and future research}
In our investigation into the development of feature extraction, the following four limitations have emerged:

\paragraph{Unexplored parameters in spectral features}
While spectral features such as MFCCs, MelSpecs, and their correlations have been widely applied in sarcasm recognition, critical aspects remain underexplored. None of the studies reviewed provide detailed insights into the specific utilization of these features. For instance, in general speaking, different MFCCs coefficients represent distinct parts of the Mel-frequency spectrum, with lower coefficients capturing broader, low-frequency patterns and higher coefficients detailing finer, high-frequency variations. We’ve known that 12–13 coefficients are used in automatic speech recognition, as they encapsulate key speech-related features, like formants. Using higher coefficients introduces more granular details but can also add noise, potentially hindering recognition if not managed carefully. In sarcasm recognition, the precise role of these selected spectral features remains unclear. Therefore, future research should focus on a more granular analysis of spectral features to better understand their relationship and contribution to sarcasm detection.

\paragraph{Absence of a best-performed speech feature set}
Linguistic research underscores the importance of speech features such as pitch, speaking rate, and intensity in conveying sarcasm. Our review confirms their frequent use in sarcasm recognition, though a best-performed feature set remains undefined. 
Given the variations in datasets, evaluation protocols, and architectures employed across the reviewed studies, it is methodologically inappropriate to draw direct conclusions linking specific features to performance. However, we encourage future research to systematically evaluate the effectiveness of the features through controlled experimental setups that can unravel the effects of features, models, and data characteristics. 

Moreover, future research should emphasize the interdependence of feature and model selection in sarcasm recognition. While commonly used features such as MFCCs, pitch, and intensity often yield positive results, their effectiveness is model-dependent. For instance, MFCCs align well with Gaussian Mixture Models (GMMs) due to their uncorrelated nature, whereas mel-filterbank features are preferred with Deep Neural Networks (DNNs), which perform better with correlated inputs. Optimizing this alignment is crucial for improving sarcasm detection accuracy in speech-based systems. 

\paragraph{Potentials of Large Language Models (LLMs) and speech-based language models}
In the audio domain, pre-trained models like WavLM \cite{Chen_2022} and SpeechLM \cite{Zhang_SpeechLM}, which build upon Wav2Vec 2.0, utilize self-supervised learning on raw speech data to capture nuanced acoustic and prosodic patterns. While Wav2Vec 2.0 emphasizes phonetic features, WavLM expands the scope by capturing long-range dependencies and prosodic dynamics such as pitch variations and speech rate, which are often associated with sarcastic expressions. Their demonstrated effectiveness in emotion and sentiment recognition \cite{Diatlova} suggests promising potential for adaptation in sarcasm recognition.

In the text modality, sarcasm often arises from the semantic incongruity, where the literal meaning of an utterance contrasts with its intended message, typically requiring contextual and world knowledge to interpret. LLMs such as ChatGPT \footnote{\url{https://chat.openai.com/}}, GPT-4 \cite{gpt4}, Claude 3 \footnote{\url{https://claude.ai/}}, and LLaMA \cite{llama} excel at modeling such nuances through pre-training on massive corpora. However, challenges remain in enhancing LLMs’ grasp of emotional tone, cultural context, and implicit social signals, which are factors critical to accurately recognizing sarcasm ~\cite{Zhang_lm}.

Moreover, recent advancements in multimodal learning have introduced transformer architectures that jointly pre-train across modalities, learning unified representations that capture cross-modal dependencies. For instance, CLIP \cite{clip} aligns visual and textual modalities within a shared embedding space, improving interpretability and semantic alignment. Extending such approaches to incorporate the audio modality allows models to jointly learn linguistic, visual, and prosodic cues, facilitating more robust and generalizable sarcasm recognition.

\paragraph{Insufficient research in the visual modality}
While DNNs, especially CNNs, have exhibited remarkable efficacy within the domain of video-based sarcasm recognition, the need for a more profound understanding of the relationship between visual features and other modality features in interpreting sarcasm remains a critical area of inquiry. Linguistic research has highlighted the interplay between audio and visual cues in sarcasm perception, indicating that sarcastic meanings are often more readily discerned through visual cues than audio cues \cite{Aguert, Rothermich}. However, further investigation is essential to unravel the complex interdependencies among cues from different modalities, which could provide valuable insights for the computational modeling of multimodal data. Additionally, improving the interpretability of these techniques is crucial for enhancing transparency and fostering continued progress in the field.

\subsection{Evolution of classification methods}
\noindent \textbf{RQ3:} \textit{As the field progresses from unimodal to multimodal sarcasm recognition, how have classification methods evolved? What are the constraints of current classification methods?}

\subsubsection{Multimodal sarcasm recognition}
For the unimodal audio-only setting, classification method simply refers to the learning algorithm chosen to classify the speech features into sarcasm or non-sarcasm. This includes basic rule-based classifiers such as Decision Trees and traditional machine learning-based classifiers like GMM, HMM, Logistic Regression, SVM, etc. Research in this area primarily focuses on identifying the most relevant sarcasm features to input into these algorithms. 

The transition to multimodal approaches has led to a leap in recognition accuracy, however, this advancement comes with increased complexity. The fusion of features from multiple modalities presents the core challenges for researchers due to their distinct characteristics, which necessitate careful consideration of how these diverse data sources are integrated. The core of multimodal sarcasm recognition lies in optimizing feature fusion strategies to enhance overall performance. Our review presents a variety of fusion strategies, ranging from foundational methods like the encoder-decoder approach to more sophisticated techniques, such as attention mechanisms. 
As the domain continues to evolve, more intricate fusion methodologies, such as quantum-based fusion, are being explored, further pushing the boundaries of sarcasm recognition technology.

Inconsistent evaluation metrics in prior studies have hindered meaningful comparisons of feature usage and classification methods.
F1-score has identified as the preferred metric as it offers a balanced assessment of recall (sarcastic samples detected) and precision (reliability of detections). Future research is suggested to involve F1-score as a metric to enhance consistency and comparability across studies.

\subsubsection{Limitations of current methods and future research}

\paragraph{Unexplored multimodal techniques}
While advanced multimodal fusion techniques have achieved significant success in sarcasm recognition, several state-of-the-art DNNs have yet to be fully explored in this domain. For instance, Graph Neural Networks (GNNs), which are designed to process graph-structured data, have shown remarkable potential in aggregating information from neighboring nodes, enabling the fusion of spatially localized features across modalities. This unique ability to capture complex relationships within graph data suggests that GNNs could be highly effective in sarcasm recognition tasks, particularly in integrating visual and audio data. Another promising yet underutilized approach is Generative Neural Networks (GenNNs), which include models such as Generative Adversarial Networks (GANs) and diffusion-based models. The primary objective of GenNNs is to generate data that closely mirrors real-world distributions, either by directly modeling these distributions or by learning transformations from simpler distributions to more complex ones. Their versatility in generating high-quality data has made them a popular choice in both unimodal and multimodal tasks, tackling challenges such as data augmentation, imputation, and fusion. In the context of sarcasm detection, where labeled data is often scarce, GenNNs could offer a powerful solution by augmenting and enhancing existing datasets.

\paragraph{Limited linguistic insights}
The review of previous multimodal research has underscored the critical role that linguistic insights play in guiding computational modeling for sarcasm recognition. There exists a consensus among the selected works that textual, audio, and visual modalities collectively contribute to the overarching sarcasm classification task. 
However, the relative effectiveness of different combinations of modality remains inconsistent between studies. For example, Castro \textit{et al}. \cite{Castro} and Ray \textit{et al}. \cite{Ray} reported that combining textual and visual modalities led to better performance compared to audio-based combinations, whereas Chauhan \textit{et al}. \cite{Chauhan_2020} and Zhang \textit{et al}. \cite{Zhang_2021} found that audio-visual combinations were more effective. These differences reflect the underlying linguistic insights. For instance, according to Camp's \cite{Camp} taxonomy of sarcasm subtypes, \textit{propositional sarcasm} involves implicit sentiment that relies on contextual interpretation; therefore, text models capable of capturing long-range dependencies perform better in this subtype. In contrast, \textit{illocutionary sarcasm} is primarily conveyed through non-verbal signals such as prosody and facial expressions, highlighting the importance of audio and visual modalities in detecting this type of sarcasm. \textit{Embedded sarcasm}, characterized by semantic incongruity, is most effective through textual features. Therefore, the integration of modalities should be guided by linguistic insights, ensuring that the models capture the nuanced ways sarcasm is expressed across speech, text, and visual cues.

An integrated linguistic-computational framework is still lacking in current research. Li \textit{et al}. \cite{Li_2022} found that native Mandarin speakers tend to rely more on visual cues in interpreting sarcastic speech, while Bromberek-Dyzman \textit{et al}. \cite{Bromberek} observed faster sarcasm recognition in audio and audio-visual modalities for bilingual participants, suggesting a facilitative role for prosody in this case. However, Deliens \textit{et al}. \cite{Deliens_2018} reported the opposite that sarcastic tone did not aid interpretation in English, and that facial expressions could even hinder it. These contradictions point to the need for cross-linguistic and cross-cultural studies that explore how different modalities interact in sarcasm perception and modeling.

\subsection{Challenges, limitations and practical implications}
\noindent \textbf{RQ4.1:} \textit{What are the primary challenges and limitations in the current development of sarcasm recognition technology?} \\
In the preceding discussion, we outlined several limitations in the development of sarcasm recognition, including data collection and annotation, feature extraction, and the techniques employed in classification models. However, at a broader level, the field continues to face challenges and limitations that need to be addressed in future research.

\subsubsection{Cross-cultural and multilingual application}
Our review identified a significant gap in cross-cultural and multilingual sarcasm detection research. Only seven out of 40 reviewed studies (17.5\%) addressed non-English languages. However, linguistic studies revealed substantial differences in sarcasm markers across cultures. For instance, in various languages, including English, French, Italian, German, and Cantonese, fundamental frequency ($F0$) (i.e., pitch) has been frequently linked to sarcasm \cite{Rockwell, Cheang_2008, Scharrer, Lœvenbruck, Cheang_2009, Anolli}. In Cantonese, Italian, and French, sarcasm is often marked by an increase in mean $F0$, whereas in German, sarcasm is signaled by a reduction in mean $F0$. Except for $F0$, reduced speech rate and lengthening of syllables or entire utterances are consistent sarcasm indicators across languages \cite{Cheang_2008, Rockwell}. These variations suggest that sarcasm is language- and culture-specific, reflecting both how speakers produce prosodic cues and how listeners interpret them. However, existing sarcasm recognition systems are predominantly trained on English data, hindering their generalizability across cultures and languages. Developing multilingual datasets offers a viable path toward building more robust systems. In addition, further research on cross-lingual transfer learning could improve the effectiveness of sarcasm recognition in broader contexts.

Another underexplored area is how multilingual speakers perceive and express sarcasm. Mandler \cite{Mandler} suggests that second language learners interpret sarcasm through the lens of their first language knowledge. For instance, Kim \textit{et al.} \cite{Kim_2014} found that Korean English speakers rely more on nonverbal cues than native Korean speakers, a trend also observed among English speakers. However, these perceptual dynamics are largely absent in current machine learning-based sarcasm detection systems. We highlight this gap as a direction for future research in multilingual speech modeling.

\subsubsection{Sarcasm recognition beyond text}
Traditionally, sarcasm has been recognized as a text-based challenge in natural language processing, typically relying on semantic cues such as contrasts between positive and negative words within a single sentence.
Numerous datasets have been developed to support these advancements, facilitating the creation of robust and generalizable models \cite{Joshi}. In contrast, sarcasm recognition in speech has not received comparable attention. Although multimodal analysis has recently revitalized interest in incorporating speech data, the field still lags behind text-based approaches. Recognizing sarcasm as a purely text-based phenomenon is outdated and inconsistent with findings from linguistic research \cite{Camp, Attardo, Li_2022, Hancock} which emphasize the multimodal nature of sarcasm.
To bridge this gap, there is a need for more comprehensive research into how different modalities interact. Such research is crucial for developing sophisticated architectures capable of recognizing sarcasm more efficiently and effectively. 

\subsubsection{Explainable and trustworthy AI}
The rapid advancements in deep learning have led to powerful AI models, yet many of these models suffer from the ``black-box" problem, where their decision-making processes are opaque and difficult to interpret. Unlike these models, which rely on vast amounts of data and complex calculations to produce predictions without revealing their inner working mechanisms, explainable AI aims to make the decision-making process transparent and comprehensible. In the context of sarcasm recognition, an explainable AI model allows us to understand how predictions are made, providing insight into the system's logic and enhancing its trustworthiness and accuracy. Research shows that transparency in AI systems fosters greater trust, particularly in critical areas such as healthcare and law enforcement. Furthermore, an interpretable model assists developers and researchers in diagnosing and refining the system by clarifying how it processes information. As sarcasm recognition research progresses, it is essential to incorporate explanations that align with linguistic theories of sarcasm cues. This focus on explainability will lead to more transparent, accountable, and reliable AI systems. \\

\noindent \textbf{RQ4.2:} \textit{What are the practical implications in real-world scenarios?}\\
Sarcasm, a commonly employed figure of speech, enables individuals to convey dissatisfaction or critique while conforming to social norms, posing a challenge in NLP. As we strive to enhance human-centered interactions with machines, the ability to recognize and understand sarcasm is crucial. Mastering sarcasm opens the door to addressing even more intricate aspects of human language, including situational irony, metaphor, hyperbole, euphemism, antithesis, and other forms of implicit expression. By tackling sarcasm effectively, we pave the way for more sophisticated and nuanced language understanding in AI systems.

In real-world deployments, the choice between unimodal and multimodal sarcasm recognition is guided by application context, resource constraints, and performance requirements. Unimodal approaches are often favored for their computational efficiency, which makes them ideal for resource-constrained environments such as social media monitoring (e.g., scanning tweets or online comments for customer sentiment) and lightweight chatbot integrations on websites to enhance customer service. However, these models struggle when sarcasm is conveyed through tone or facial expressions. In contrast, multimodal approaches that fuses textual, audio, and visual cues, offer higher systems accuracy by capturing enriched cross-modal relationships. For example, a voice assistant (e.g., Alexa, Google Assistant) equipped with sarcasm detection can use both words and tone to avoid misinterpreting a frustrated ``sure, that's just perfect'' as genuine praise. In clinical context, multimodal models can help detect sarcasm or distress more reliably by combining speech with facial expressions. Video content moderators on platforms like YouTube or TikTok can deploy multimodal systems to catch sarcasm that text alone would miss, supporting the identification of hate speech or misinformation effectively. Moreover, individuals with neurodegenerative conditions often face challenges interpreting non-literal language. By highlighting or translating sarcastic remarks into their literal intent, these systems can facilitate social integration for these users. However, multimodal systems often have high computation demands, relying on advances in high-performance hardware. Ultimately, practitioners should balance the trade-off between efficiency and accuracy in light of the specific goals of deployment when developing the system.

\section{Conclusion}
This systematic review charts the evolution of speech-based sarcasm recognition, revealing how the field has progressed from simple audio analysis to sophisticated multimodal approaches. Through analysis of 40 papers, we find that while traditional prosodic and spectral features remain fundamental, deep learning architectures, particularly those leveraging attention mechanisms, have transformed how we detect sarcasm in speech. Our findings suggest that no single modality or feature set is sufficient; rather, it is the strategic fusion of acoustic, textual, and visual cues that yields the most robust results. This is evident in the success of attention-based architectures that can capture subtle cross-modal relationships. However, significant challenges remain: current datasets are limited in size and spontaneity, prosodic feature sets lack standardization, and cross-cultural aspects of sarcasm remain underexplored. By bridging linguistics and computational approaches, this review provides a foundation for researchers tackling these challenges while highlighting promising directions for future work.

In conclusion, this review reveals that while deep learning and multimodal approaches have advanced sarcasm recognition, critical challenges persist. The field stands at an intersection where computational advances in attention mechanisms and feature fusion meet linguistic insights about how sarcasm manifests across modalities and cultures. Future progress requires three key developments: first, larger, more diverse datasets that capture spontaneous sarcastic speech across languages; second, standardized prosodic feature sets informed by linguistic research on sarcasm markers; and third, more sophisticated fusion architectures that can better model the subtle interplay between acoustic, textual, and visual cues. Success in these areas would not only advance our theoretical understanding of sarcasm but also enable practical applications that could significantly impact HMI and assist individuals with pragmatic language difficulties. As sarcasm recognition continues to evolve, maintaining this bridge between linguistic theory and computational practice will be crucial for developing systems that can understand and respond to this nuanced aspect of human communication.

\bibliographystyle{IEEEtran}
\bibliography{references}

\begin{IEEEbiography}[{\includegraphics[width=1in,height=1.25in,clip,keepaspectratio]{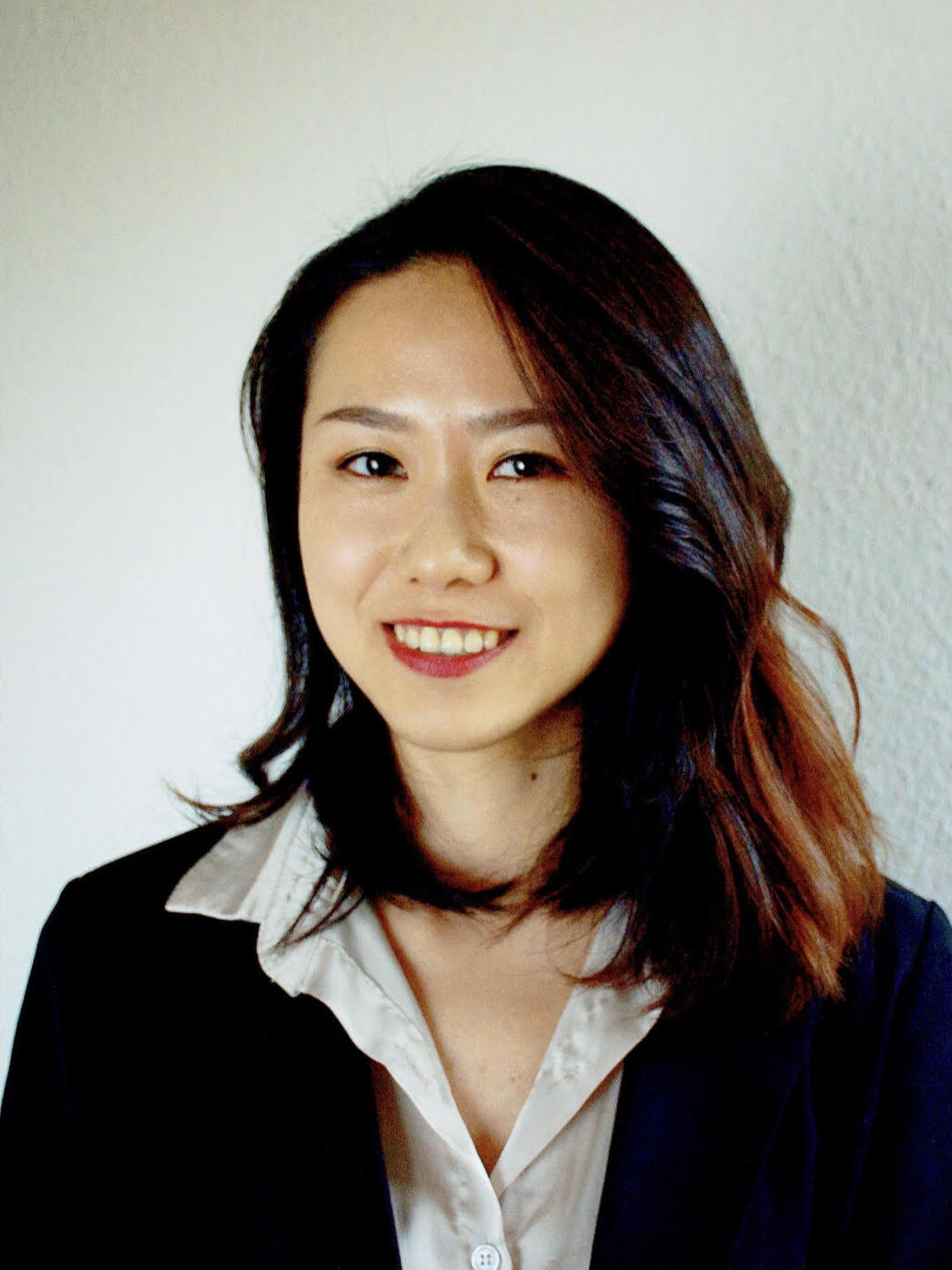}}]
{Xiyuan Gao} received the Master's degree in Speech and Language Processing from Konstanz University, Konstanz, Germany. She is currently a Ph.D. candidate from Speech Technology at the Faculty Campus Fryslân, University of Groningen, the Netherlands. Her research interests include linguistic insights driven sarcasm detection, multi-modal framework, speech technology.
\end{IEEEbiography}

\begin{IEEEbiography}[
{\includegraphics[width=1in,height=1.25in,clip]{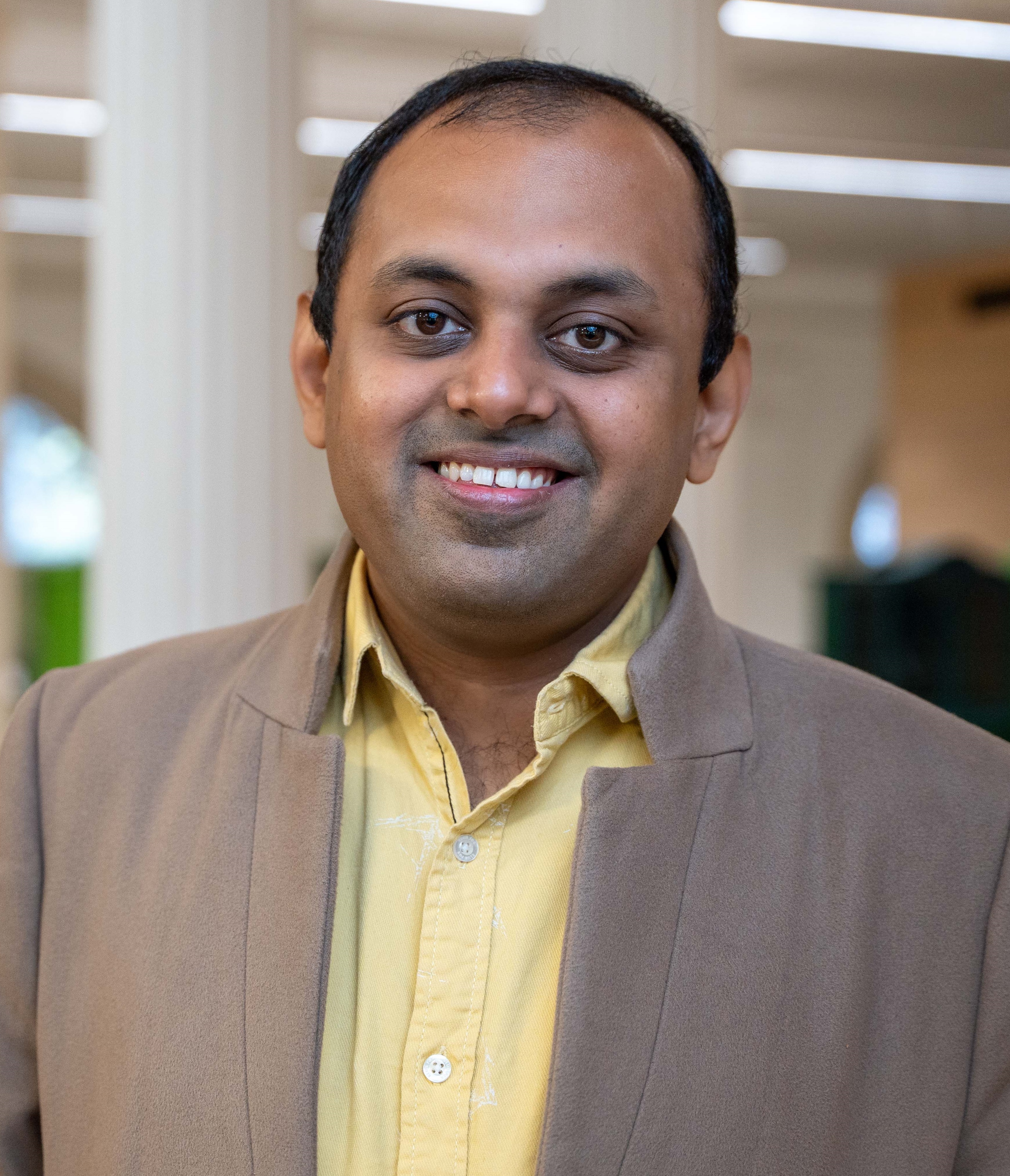}}
]
{Shekhar Nayak} received the Ph.D. degree and the M.Tech. degree from the Indian Institute of Technology (IIT) Hyderabad and IIT Delhi, India, in 2019 and 2011, respectively. He served as a Technology Consultant from 2011-2013 at Hewlett Packard Enterprise and as a Senior Chief Engineer at Samsung R\&D Institute, Bangalore, India from 2019-2021 in the Voice Services Department. He was a Research Assistant at the Institute for Infocomm Research (I2R), Agency for Science, Technology and Research (A*STAR), Singapore in 2015. Dr. Nayak is currently an Assistant Professor of Speech Technology at the Faculty Campus Fryslân, University of Groningen, the Netherlands since 2021. His research interests include speech recognition, speech synthesis and multi-modal signal processing and deep learning.
\end{IEEEbiography}

\begin{IEEEbiography}[
{\includegraphics[width=1in,height=1.25in,clip,keepaspectratio]{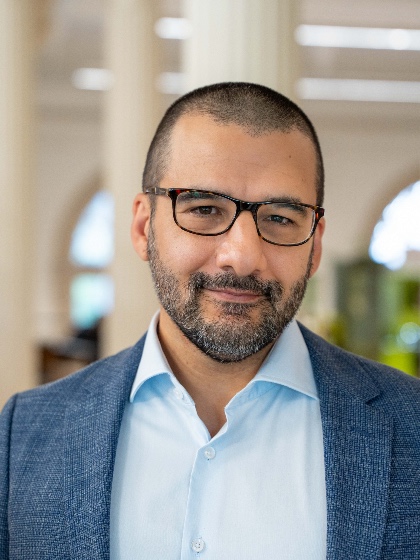}}
]
{Matt Coler} received the Ph.D. degree from the Free University of Amsterdam, the Netherlands, in 2010. After a postdoctoral appointment at his alma mater, he joined an AI start-up focusing on acoustic sensors, where he served as Head of the Cognitive Systems Unit. In 2012, he returned to academia. Dr. Coler is currently an Associate Professor of Speech Technology at the Faculty Campus Fryslân, University of Groningen, the Netherlands, where he works as the Director of the M.Sc. Voice Technology program and the Chair of the faculty Research Institute. His research interests include topics around speech technology with small data, low-resource languages, pragmatics in speech technology, and ethics. 
\end{IEEEbiography}

\clearpage
\setcounter{page}{1}

\renewcommand{\thetable}{A\arabic{table}}
\setcounter{table}{0}

\onecolumn
\noindent{\textbf{Appendix A: Supplementary tables}}
\begin{table}[H]
\footnotesize
\centering
\caption{Summary of multimodal sarcasm recognition. \textit{Languages: en = English, es = Spanish, zh = Chinese, hi = Hindi, it = Italian kn = Kannada. Modality: T = text, A = audio, V = visual.}} \label{tablemultimodal}
\begin{tabularx}{\textwidth}{
>{\raggedright\arraybackslash}p{4cm} 
>{\raggedright\arraybackslash}p{2.0cm} 
>{\raggedright\arraybackslash}p{1.8cm} 
>{\raggedright\arraybackslash}p{2.5cm} 
>{\raggedright\arraybackslash}p{2.6cm}
>{\raggedright\arraybackslash}X
}
\toprule
\textbf{Specific Features} & \textbf{Fusion Method} & \textbf{Classifier} & \textbf{Dataset} & \textbf{Evaluation} & \textbf{Best Performance(\%)} \\
\midrule
A \& T: unknown\newline T: BiGRU embeddings &
Attention mechanism & 
Softmax layer & 
SEEmoji MUStARD (en)&
5-fold cross-validation & 
F1-score: 80.0 \cite{Chauhan_2022} \\

A: MFCCs, MelSpec, SC, delta\newline T: BERT embeddings\newline V: ResNet-152 embeddings &
Encoder-decoder  &
Softmax layer & 
MUStARD (en) &
5-fold cross-validation & 
F1-score: 79.0 \cite{Sun} \\

A: MFCCs, MelSpec, ZCR, SC\newline T: BERT embeddings\newline V: ResNet embeddings &
Attention mechanism &
Softmax layer & 
MUStARD (en) &
5-fold cross-validation &
F1-score: 75.2 \cite{Li_2024} \\

A: MFCCs, ZCR, chroma\newline T: BERT embeddings\newline V: ResNet-152 embeddings &
Encoder-decoder & 
FCNN & 
MUStARD (en) &
5-fold cross-validation & 
F1-score: 75.01 \cite{Ding} \\

A: VGGish embeddings\newline T: BiGRU embeddings\newline V: EfficientNet embeddings &
Attention mechanism  &
Softmax layer & 
MUStARD (en) &
5-fold cross-validation & 
F1-score: 74.6 \cite{Zhang_2023a} \\

A: MFCCs, ZCR\newline T: BERT embeddings\newline V: ResNet-50 embeddings  &
Attention mechanism & 
Softmax layer & 
MUStARD (en) &
5-fold cross-validation & 
F1-score: 74.5 \cite{Wu} \\

A: MFCCs, MelSpec, SC, delta\newline T: BiGRU embeddings\newline V: ResNet-152 embeddings & 
Attention mechanism & 
Softmax layer & 
{MUStARD} (en) & 
5-fold cross-validation & 
F1-score: 72.57 \cite{Chauhan_2020} \\

A: MFCCs, MelSpec, SC, delta\newline T: BERT embeddings\newline V: ResNet-152 embeddings & 
Encoder-decoder & 
SVM & 
{MUStARD} (en) & 
5-fold cross-validation & 
F1-score: 71.6 \cite{Castro} \\

A: MFCCs\newline T: tokenized sequences &
Encoder-decoder & 
SVM & 
MUStARD (en) &
5-fold cross-validation & 
F1-score: 70.35 \cite{Bharti} \\

A: VGGish embeddings\newline T: BERT embeddings\newline V: EfficientNet embeddings &
Quantum-based  &
Softmax layer & 
MUStARD (en) &
Train-test split & 
F1-score: 83.7 \cite{Liu} \\

A: Wav2Vec2.0\newline T: BART embeddings\newline V: ResNet-152 embeddings &
Attention mechanism, Collaborative gating &
Softmax layer & 
MUStARD (en) &
Train-test split &
F1-score: 79.71 \cite{Tomar} \\

A: MFCCs, MelSpec, SC\newline T: BERT embeddings\newline V: RagNet embeddings &
Encoder-decoder   &
FCNN & 
MUStARD (en) &
Train-test split &
F1-score: 76.7 \cite{Pandey} \\

A: VGGish embeddings\newline T: BERT embeddings\newline V: ResNet-152 embeddings &
Attention mechanism  &
Softmax layer & 
MUStARD (en) &
Train-test split &
F1-score: 76.36 \cite{Zhang_2023b} \\

A: Wav2Vec2.0\newline T: BART embeddings\newline V: ResNet-152 embeddings &
Attention mechanism &
Softmax layer & 
MUStARD (en) &
Train-test split &
F1-score: 75.57 \cite{Zhang_2024} \\

A: MFCCs, pitch tracking VUV, peak slope, MDQ\newline T: BERT embeddings\newline V: ResNet-152 embeddings  &
Attention mechanism & 
Softmax layer & 
MUStARD (en) &
Train-test split & 
F1-score: 73.97 \cite{Zhang_2021} \\

A: VGGish embeddings\newline T: Albert embeddings\newline
V: EfficientNet embeddings &
Attention mechanism &
Softmax layer & 
MUStARD (en) &
Train-test split &
F1-score: 68.88 \cite{Tiwari} \\

A: MFCCs\newline T: TF-IDF features\newline V: LGIP features &
Encoder-decoder &
Sigmoid layer & 
MUStARD (en) &
Train-test split &
Accuracy: 93.6 \cite{Krishnamaneni} \\

A: MFCCs, pitch, glottal, harmonic and phase distortions, formant\newline T: Albert embeddings\newline V: FAU, (non-) rigid facial shape &
Attention mechanism & 
Softmax layer & 
MUStARD (en) &
Train-test split & 
Accuracy: 79.41 \cite{Hasan} \\

A: MFCCs, VUV, glottal\newline T: BERT embeddings\newline V: ResNet-101 embeddings  &
Attention mechanism & 
FCNN & 
MUStARD (en) &
Train-test split & 
Accuracy: 76.82 \cite{Pramanick} \\

\bottomrule
\end{tabularx}
\end{table}

\begin{table*}[t]
\footnotesize
\centering
\begin{threeparttable}
\begin{tabularx}{\textwidth}{
>{\raggedright\arraybackslash}p{4.2cm} 
>{\raggedright\arraybackslash}p{2.0cm} 
>{\raggedright\arraybackslash}p{1.8cm} 
>{\raggedright\arraybackslash}p{2.5cm} 
>{\raggedright\arraybackslash}p{2.6cm}
>{\raggedright\arraybackslash}X
}
\toprule

A: MFCCs, MelSpec, SC, delta\newline T: BiGRU embeddings &
Encoder-decoder &
Sigmoid layer & 
MUStARD (en) &
Unknown &
F1-score: 74.67 \cite{Azahouani} \\

A: ComParE feature set\newline T: BERT embeddings\newline Emotion: Wav2cev2.0 embeddings\newline Sentiment: SiEBERT embeddings &
Attention mechanism &
Softmax layer & 
MUStARD++ (en) &
Train-test split &
F1-score: 74.3 \cite{Gao_2024} \\

A: MFCCs, MelSpec, prosodics\newline T: BERT embeddings\newline V: ResNet-152 embeddings &
Collaborative gating &
Sigmoid layer & 
MUStARD++ (en) &
Unknown evaluation & 
F1-score: 74.2 \cite{Ray} \\

A: MFCCs\newline T: FastText embeddings  &
Attention mechanism & 
Sigmoid layer & 
MASAC (hi) &
Train-test split & 
F1-score: 71.1 \cite{Bedi} \\

A: MFCCs, ZCR, SC, bandwidth, pitches, chroma, tonnetz\newline T: GloVe embeddings\newline V: ResNet-152 embeddings & 
Encoder-decoder & 
SVM & 
Spanish dataset (es) & 
Train-test split & 
Accuracy: 93.1 \cite{Alnajjar} \\

A: VGGish embeddings\newline T: BERT embeddings\newline V: ResNet embeddings &
Attention mechanism &
FCNN & 
CMMA (zh) &
Train-test split &
F1-score: 75.72 \cite{Zhang_CMMA} \\

A: MFCCs, PLP, pitch, formant, duration, pauses, syllables per second, energy, HNR\newline T: CNN embeddings &
Encoder-decoder &
Sigmoid layer & 
Custom dataset (en) &
5-fold cross-validation & 
F1-score: 73.9 \cite{Gent} \\

A: MFCCs, MelSpec, SC\newline T: BERT embeddings\newline V: ImageNet embeddings &
Encoder-decoder &
FCNN & 
Custom dataset (en) &
5-fold cross-validation &
Accuracy: 99.05 \cite{Murthy} \\

A: glottal, jitter, shimmer, HNR, energy\newline T: sentiment polarity\newline V: facial and eye aspect ratio &
Encoder-decoder & 
FCNN & 
Custom dataset (en) &
Train-test split & 
Accuracy: 81.8 \cite{Hiremath} \\

A: eGeMAPS feature set \cite{GeMAPS}\newline V: FAU, eye gaze &
Encoder-decoder &
Several models\tnote{a} & 
Custom dataset (it) &
4-fold cross-validation &
F1-score: 80 \cite{Spitale} \\

A: duration, pitch, intensity\newline Ar: measurements of tongue \& lip movements&
Encoder-decoder &
Random Forest &
Custom dataset (zh) &
Train-test split &
Accuracy: 74.1 \cite{Geng} \\

A: MFCCs, pitch, intensity\newline T: n-grams, sentiment scores, lexical patterns &
Encoder-decoder &
FCNN & 
Custom datsset (kn) &
Train-test split &
F1-score: 76.5 \cite{Manohar} \\

\bottomrule
\end{tabularx}
\begin{tablenotes}
\scriptsize
\item[a] Several Models include Logistic Regression, SVM, Random Forest, XGBoost, AdaBoost, Decision Tree, and Multilayer Perceptron. The reported performance represents the average result across all models.
\end{tablenotes}
\end{threeparttable}
\end{table*}

\begin{table*}[h]
\footnotesize
\centering
\caption{Summary of unimodal (audio-only) sarcasm recognition.}
\begin{tabularx}{\textwidth}{
  >{\raggedright\arraybackslash}p{1.6cm} 
  >{\raggedright\arraybackslash}p{3.5cm} 
  >{\raggedright\arraybackslash}p{2.5cm} 
  >{\raggedright\arraybackslash}p{2.5cm} 
  >{\raggedright\arraybackslash}p{2.8cm} 
  >{\raggedright\arraybackslash}p{2.8cm} 
}
\toprule
\textbf{Feature Type} & \textbf{Specific Features} & \textbf{Classifier} & \textbf{Dataset} & \textbf{Evaluation} & \textbf{Best Performance} \\
\midrule
Spectral, contextual & MFCCs, deltas, acceleration coefficients; pitch, energy; context (e.g., laughter, pause, gender) & Decision
Trees & Switchboard \cite{switchboard} and Fisher \cite{fisher} (en) & Leave-one-out cross-validation & F1-score: 70\% \cite{Tepperman} \\

Spectral, prosodic, voice quality & MFCCs, PLP; pitch contours; harmonicity, frequencies and bandwidths of first three formants & Gaussian Mixture Models, and a simple perceptron & COST 2102 Italian Database of Emotional Speech (it) & Leave-one-speaker-out & Accuracy: 58\% \cite{Atassi_2010} \\

Prosodic & Pitch, intensity, speaking rate & Logistic Regression & Custom dataset (en) & 10-fold cross-validation & Accuracy: 81.57\% \cite{Rakov} \\

Spectral, prosodic & MFCCs; pitch variations & Hidden Markov Model & Custom dataset (hi) & Average 10 audio files & Accuracy: 70 \cite{Mathur} \\

Spectral, prosodic, voice quality & ComParE feature set & Support Vector Machine & Custom dataset (de) & 5-fold cross-validation & Recall: 69.6 \cite{Burkhardt} \\

Spectral, prosodic & MFCCs, the log of filterbank energy; pitch & Random Forest  & Custom dataset (hi) & Train-test split & F1-score: 68\% \cite{Arun} \\

Spectral & MelSpec & Conventional Neutral Network & MUStARD (en) & 5-fold cross-validation
 & F1-score: 72\% \cite{Gao_2022} \\

Spectral, prosodic & MelSpec, MFCCs, chroma, SCt, SR, SBW, tonnetz, ZCR; intensity & Support Vector Machine & IIT-KGP-SEHSC (hi) & Train-test split & F1-score: 73\% \cite{Sacheth} \\

\bottomrule
\end{tabularx}
\label{audio_features_unimodal}
\end{table*}

\end{document}